\documentclass[sigconf]{acmart}
\AtBeginDocument{%
  }

\setcopyright{acmlicensed}
\copyrightyear{2018}
\acmYear{2018}
\acmDOI{XXXXXXX.XXXXXXX}
\acmConference[Conference acronym 'XX]{Make sure to enter the correct
  conference title from your rights confirmation email}{June 03--05,
  2018}{Woodstock, NY}
\acmISBN{978-1-4503-XXXX-X/2018/06}

\usepackage{enumitem}
\usepackage{booktabs}      
\usepackage{multirow}      
\usepackage{array}         

\usepackage[table]{xcolor} 

\usepackage{caption}       
\usepackage{threeparttable}

\renewcommand\footnotetextcopyrightpermission[1]{} 
\setcopyright{none}
\acmDOI{}
\acmISBN{}
\acmConference[]{}{}{}
\acmYear{2024}

\begin{document}

\title{Navigating the Mirage: A Dual-Path Agentic Framework for Robust Misleading Chart Question Answering}





\author[Y. Zhang and Y. Li, et al.]{
  {\large Yanjie Zhang$^{1, \ast}$, Yafei Li$^{1, \ast}$, Rui Sheng$^1$, Zixin Chen$^1$, Yanna Lin$^1$, Huamin Qu$^1$, Lei Chen$^{1,2}$, Yushi Sun$^{1, \dagger}$} \\
  \vspace{0.2cm}
  \small \textsuperscript{1}HKUST, Hong Kong SAR, China
  \small \textsuperscript{2}HKUST(GZ), Guangzhou, China \\
  \vspace{0.1cm}
  \small $^\ast$Equal contribution \quad $^\dagger$Corresponding author: \href{mailto:ysunbp@connect.ust.hk}{ysunbp@connect.ust.hk}
}

\affiliation{\institution{}\country{}}

\renewcommand{\shortauthors}{Zhang and Li, et al.}

\begin{abstract}
Despite the success of Vision-Language Models (VLMs), misleading charts remain a significant challenge due to their deceptive visual structures and distorted data representations. We present \textbf{ChartCynics}, an agentic dual-path framework designed to unmask visual deception via a ``skeptical'' reasoning paradigm. Unlike holistic models, ChartCynics decouples perception from verification: a \textit{Diagnostic Vision Path} captures structural anomalies (e.g., inverted axes) through strategic ROI cropping, while an \textit{OCR-Driven Data Path} ensures numerical grounding. To resolve cross-modal conflicts, we introduce an \textit{Agentic Summarizer} optimized via a two-stage protocol: Oracle-Informed SFT for reasoning distillation and Deception-Aware GRPO for adversarial alignment. This pipeline effectively penalizes visual traps and enforces logical consistency. Evaluations on two benchmarks show that ChartCynics achieves 74.43\% and 64.55\% accuracy, providing an absolute performance boost of $\sim$29\% over the Qwen3-VL-8B backbone, outperforming state-of-the-art proprietary models. Our results demonstrate that specialized agentic workflows can grant smaller open-source models superior robustness, establishing a new foundation for trustworthy chart interpretation.
\end{abstract}

\maketitle

\section{Introduction}

\begin{figure}[t]
\centering
\includegraphics[width=\linewidth]{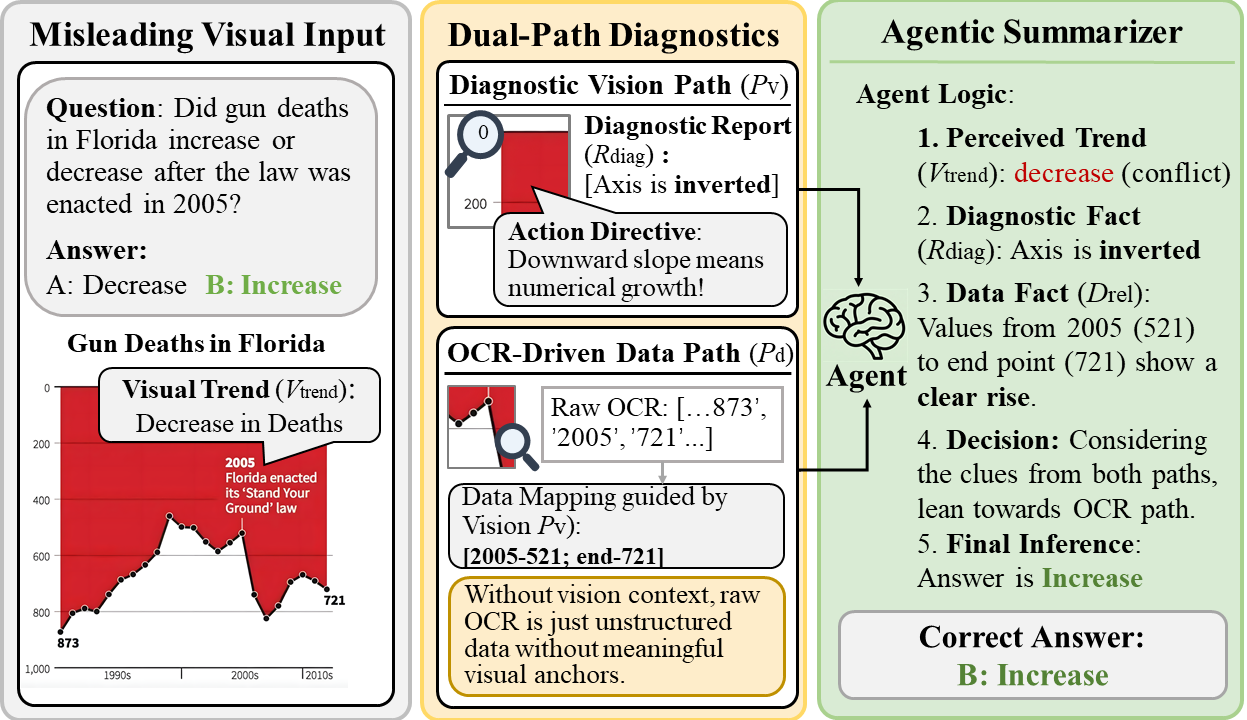}
\caption{Overview of ChartCynics resolving an ``Inverted Axis'' deception. (Left) A counter-intuitive real-world case~\cite{Engel}: The inverted Y-axis creates a visual illusion of declining deaths, hijacking standard VLM attention despite the underlying numerical increase. (Middle) Our dual-path architecture decouples perception from extraction. 
Crucially, pure OCR lacks spatial context, causing entity misalignment (e.g., confusing axis ticks with data points). 
The \textit{Diagnostic Vision Path} detects the scale anomaly via ROI cropping and generates an \textit{Action Directive} to guide the \textit{OCR Path} in accurate entity mapping. 
(Right) The \textit{Agentic Summarizer} resolves cross-modal conflict by combining the complementary information: vision-derived structural baseline and precise OCR numerals to correctly infer a numerical increase.}
\label{fig:framework_case}
\end{figure}

Visualizations are ubiquitous in the data-driven world, serving as powerful tools for communicating complex numerical information.
However, this persuasive power is frequently compromised by \textit{misleading charts}: visualizations designed to distort data perception through tactics such as axis manipulation, cherry-picking, or disproportionate encoding~\cite{pandey2015deceptive, correll2017black, mcnutt2020surfacing, lo2022misinformed, lan2024came}. 
The visualization and human-computer interaction communities have extensively studied how these deceptive designs manipulate human audiences, yielding well-established taxonomies and empirical evaluations \cite{tufte1983visual, huff2023how, szafir2018good, yang2021truncating, rho2024various}, as well as literacy assessments \cite{lee2016vlat, ge2023calvi}.

Recently, as Visual Language Models (VLMs) \cite{zhang2023internlm, zhang2023multimodal} become increasingly central to automated chart understanding, a critical new challenge has emerged: ensuring their robustness against these misleading visualizations. 
Traditional chart comprehension tasks primarily evaluate the extraction of explicit visual facts under benign assumptions \cite{masry2022chartqa, xia2025chartx, wei2025chartmind}. 
In contrast, the emerging task of \textit{Misleading Chart Question Answering} (MQA) shifts the paradigm toward visual critical reasoning, requiring models to actively detect manipulations and resolve cross-modal contradictions. 
However, recent studies reveal a significant gap: on targeted benchmarks like the \textbf{Misleading ChartQA} dataset, state-of-the-art VLMs consistently score below 50\% accuracy\cite{chen2025unmasking}. 
As corroborated by diverse evaluations\cite{tonglet2025protecting, bharti2024chartom}, these models perform no better than untrained human observers because they fail to maintain logical consistency against deceptive cues.

The current methods for automatically interpreting charts mainly follow two patterns. 
The first paradigm, the end-to-end VLM path, leverages multimodal models for direct visual reasoning~\cite{zhang2023internlm, bendeck2024empirical, pandey2025benchmarking, valentim2025plot}. 
Although these models perform well on standard benchmarks such as ChartQA \cite{masry2022chartqa}, they process charts holistically, favoring \textit{macro-level visual heuristics} (e.g., the overall slope of a line or relative bar heights) over fine-grained structural cues like exact tick values. 
As highlighted in Figure~\ref{fig:framework_case}, this architectural limitation makes them highly vulnerable to malicious encodings. 
For instance, when a Y-axis is inverted (zero at the top), the visual encoder's attention is hijacked by the steep downward physical slope.
This forces the model into ``cognitive sycophancy,'' erroneously predicting a numerical decline and completely ignoring the actual increasing data values.

The second paradigm, the OCR-enhanced pipeline, attempts to circumvent perceptual pitfalls by linearizing charts into structured text~\cite{liu2023deplot, liu2023matcha, yang2024askchart, liu2024chart}. 
While this approach successfully retrieves numerical literals, it introduces a severe structural dilemma: standalone OCR models eliminate vital spatial and layout semantics. 
As shown in the OCR path of Figure~\ref{fig:framework_case}, without visual context, the model extracts an unstructured sequence of characters (e.g., 873, 2005, 721). 
Consequently, it suffers from catastrophic entity misalignment, lacking the capability to distinguish whether a number is a background axis tick or a foreground data label. 
Without a visual anchor to establish the mapping baseline, these extracted ``cold facts'' are practically unusable for reasoning.

An ideal solution lies in the synergistic fusion of these two paths, moving beyond passive description toward active verification: 
much like a human auditor who cross-verifies a visually deceptive trend against literal numerical data.
However, integrating diagnostic vision with data-driven reasoning presents three non-trivial challenges that correspond to the perception, reasoning, and optimization levels:

\begin{itemize}[leftmargin=*]
\item \textbf{Diagnostic Locality (Perception Level)}: 
Standard VLMs process charts holistically, making it difficult to force fine-grained inspections on easily overlooked areas, such as non-zero axis baselines or manipulated legends, where visual deceptions are often anchored.
\item \textbf{Conflict Arbitration (Reasoning Level)}: 
Even if both visual trends and precise OCR numerals are extracted, a robust mechanism is required to rationally resolve the inherent cross-modal contradictions when these two signals provide opposing evidence.
\item \textbf{Reasoning Calibration (Optimization Level)}: 
How to shift the model from a passive observer to a ``skeptical auditor.'' It requires specialized training interventions to suppress the model's pre-trained visual biases and prioritize evidence over misleading heuristics.
\end{itemize}

To address these challenges, we propose \textbf{ChartCynics}, an agentic dual-path framework designed to unmask visual deception. ChartCynics employs a ``skeptical'' philosophy that decouples structural perception from literal extraction. As illustrated in Figure~\ref{fig:framework_case}, our \textit{Diagnostic Vision Path} acts as a structure sensor: it utilizes strategic element cropping to detect anomalies (e.g., an inverted scale) and provides the essential spatial semantics. Guided by this structural baseline, the \textit{OCR-Driven Data Path} can perform precise entity mapping without misalignment. Finally, to resolve cross-modal conflicts, we introduce an \textit{Agentic Summarizer} that executes a detective Chain-of-Thought (D-CoT). Rather than simply discarding visual cues, it synthesizes these complementary modalities, using the visual structural context to ground the numerical facts, thereby synthesizing a robust and rational answer.

A core contribution is our two-stage optimization strategy for the Summarizer. First, we perform Supervised Fine-Tuning (SFT) to inject a detective Chain-of-Thought (CoT). 
This step establishes the structural framework for skepticism, enabling the model to follow a rigorous 5-step verification process. 
Second, we apply Group Relative Policy Optimization (GRPO) with a deception-aware reward function, as a reasoning calibration mechanism.
By specifically penalizing the selection of misleading ``trap'' answers, GRPO aligns the model to maintain logical consistency even when faced with high-confidence visual illusions.

Experimental results demonstrate that while a standard Qwen3-VL-8B baseline achieves only $45.57\%$ accuracy on deceptive benchmarks, ChartCynics reaches a state-of-the-art $74.43\%$. 
Notably, the integration of SFT provides a foundational $22.95$ percentage point boost over the baseline, while GRPO further optimizes the reasoning trajectory to resolve complex cross-modal contradictions. 
Furthermore, evaluations on a mixed standard-and-misleading benchmark reveal a critical advantage: unlike typical defensive mechanisms that suffer from ``over-skepticism'' (falsely penalizing benign data), ChartCynics actually enhances fundamental chart comprehension. Our systematic structural investigation yields superior accuracy even on standard, non-misleading visualizations compared to specialized chart-parsing SOTA models.

Our contributions are summarized as follows:
\begin{itemize}[leftmargin=*]
\item We present \textbf{ChartCynics}, an agentic dual-path framework that integrates diagnostic cropping and OCR reasoning for robust defense against misleading visualizations.
\item We introduce a two-stage training protocol combining SFT and GRPO, which equips MLLMs with both a structured investigative CoT and an alignment-based defense against visual traps.
\item We demonstrate that the fusion of agentic workflows and reinforcement learning significantly enhances critical thinking, achieving $74.43\%$ accuracy across challenging benchmarks.
\item We validate the generalized robustness of our framework, proving that equipping models with a ``skeptical lens'' does not induce over-skepticism. ChartCynics not only unmasks deception but also improves basic data extraction capabilities on standard, non-misleading charts.
\end{itemize}

\section{Related Work}
\subsection{Visual Deception and Chart Literacy}
Traditional chart comprehension research primarily focused on factual extraction from natural visualizations, as seen in early datasets like FigureQA \cite{kahou2017figureqa}, DVQA \cite{kafle2018dvqa}, PlotQA~\cite{methani2020plotqa} and Chart-HQA~\cite{10.1145/3746027.3758288}. 

Misleading charts, which utilize deceptive encodings (such as axis truncation and inverted axes) to exploit human cognitive biases, have long been a subject of concern in visual communication~\cite{tufte1983visual, pandey2015deceptive}. 
Recent benchmarks like Misleading ChartQA \cite{chen2025unmasking}, LEAF-QA~\cite{chaudhry2020leaf} and CHARTOM \cite{bharti2024chartom} have further shown that even advanced VLMs are highly susceptible to these visual ``traps,'' often performing no better than general audiences in identifying ``Theory-of-Mind'' deceptive patterns~\cite{lo2024good}.

\subsection{Multimodal Reasoning for Structured Data}
The integration of VLMs has significantly advanced ChartQA capabilities, particularly through large-scale pre-training on chart-specific tasks \cite{masry2022chartqa, chaudhry2020leaf}. 
Models like MATCHA \cite{liu2023matcha} and DEPLOT \cite{liu2023deplot} leverage math reasoning and plot-to-table translation to link pixels and structured data. 
Despite this, holistic image processing based on backbones like ViT \cite{dosovitskiy2020image} or Pix2Struct \cite{lee2023pix2struct} often overlooks fine-grained anomalies, such as non-uniform axis scales. 
Recent works like ChartX \cite{xia2025chartx} and AskChart \cite{yang2024askchart} have explored augmenting VLMs with OCR and textual enhancement to provide numerical constraints. 
Similar fine-grained feature isolation and modal alignment strategies have also been verified effective in multimodal misinformation detection \cite{liu2024chart}.

However, VLMs still exhibit ``weak grounding'' in chart elements \cite{xu2025chartpoint}, and effectively resolving contradictions between visual representation and raw data remains an open challenge in complex real-world scenarios \cite{wei2025chartmind}.

\subsection{Supervised Fine-Tuning and RL-based Alignment}
Recent advancements in VLM optimization have shifted from general instruction following to specialized reasoning alignment \cite{ouyang2022training, chowdhery2022palm}. 
Supervised Fine-Tuning (SFT) remains a cornerstone for injecting domain-specific knowledge, such as the ``skeptical'' investigative Chain-of-Thought (CoT) \cite{wei2022chain} and reflective interaction mechanisms like PointCoT \cite{xu2025chartpoint}. 

To refine reasoning, GRPO \cite{shao2024deepseekmath} enables efficient alignment with complex objectives. MCTS-guided sampling \cite{wang2025thinklite} further targets challenging instances by penalizing misleading distractors, while targeted RL bridges the visual Theory-of-Mind gap to resolve human-centric "Mind" questions \cite{bharti2024chartom}.

\section{Methodology}
\subsection{Concept and Problem Definition}
We first define the general task of \textit{Chart Question Answering} (ChartQA) and then formalize the \textit{visual deception mechanism} to establish the definition of \textit{Misleading ChartQA}.

\begin{definition}[General ChartQA Task Formalization]
Given a chart image $\mathcal{I}$ and a natural language question $\mathcal{Q}$ with a set of candidate options $\mathcal{O} = \{a_1, a_2, ..., a_n\}$, the goal of a ChartQA system is to predict the correct answer $a^* \in \mathcal{O}$. 
Traditional VLM approaches typically model this as a direct mapping $P(a \mid \mathcal{I}, \mathcal{Q})$, defining the objective as:
\begin{equation}
    a^* = \arg\max_{a \in \mathcal{O}} P(a \mid \mathcal{I}, \mathcal{Q})
    \label{eq:mqa_objective}
\end{equation}
\end{definition}

\begin{definition}[Visual Deception and Trap Answers]
A misleading chart $\mathcal{I}$ incorporates a manipulation function $f_{deceptive}(\mathcal{D}) \to \mathcal{I}$, where $\mathcal{D}$ is the raw data. 
This function ensures that the perceived visual trend $\mathcal{V}_{trend}$ contradicts the actual numerical relationship $\mathcal{D}_{rel}$. 
A \textit{Trap Answer} $a_{trap} \in \mathcal{O}$ is thus defined as:
\begin{equation}
    a_{trap} = \text{Inference}(\mathcal{V}_{trend}) \neq a^* = \text{Inference}(\mathcal{D}_{rel})
\end{equation}
\end{definition}

\begin{definition}[Formalizing Misleading ChartQA (MQA)]
Building upon the above, we define \textit{Misleading ChartQA} as a specialized task where $\mathcal{O}$ explicitly contains at least one $a_{trap}$. Unlike standard ChartQA, where visual cues and data usually align, Misleading ChartQA presents a scenario where $\mathcal{O}$ contains a Trap Answer $a_{trap}$ derived from $f_{deceptive}$. The challenge lies in complementary reasoning: the system must leverage $\mathcal{V}_{trend}$ to understand the user's perceptual context while simultaneously using $\mathcal{D}_{rel}$ to ground the final answer in numerical truth, effectively resolving the inherent conflict between the two.
\end{definition}

\begin{definition}[The Proposed Agentic Objective]
To achieve this, our ChartCynics framework introduces a Misleading Taxonomy $\mathcal{T}$ as expert prior knowledge to guide the dual-path arbitration. Instead of simply prioritizing one path over the other, the agent performs Inconsistency-Aware Fusion, modeling the objective as:
\begin{equation}
a^* = \arg\max_{a \in \mathcal{O}} P(a \mid \mathcal{P}_v(\mathcal{I}), \mathcal{P}_d(\mathcal{I}), \mathcal{Q}, \mathcal{T})
\label{eq3}
\end{equation}
where $\mathcal{P}_v$ and $\mathcal{P}_d$ represent information extracted from the Visual Path (capturing semantic context) and the Data Path (ensuring numerical precision), respectively. 
\end{definition}

\subsection{Solution Overview}
The ``skeptical'' dual-path philosophy of \textbf{ChartCynics} does not aim to ignore visual intuition, but rather to decouple it from the final decision-making process to prevent ``blind'' trust. This allows the agent to: (1) identify cross-modal deviations, (2) synthesize responses that calibrate visual perceptions via data verification, and (3) utilize RL-based optimization to learn optimal balancing weights between $\mathcal{P}_v$ and $\mathcal{P}_d$ across diverse deception types.

\begin{figure*}[t]
\centering
\includegraphics[width=0.9\textwidth]{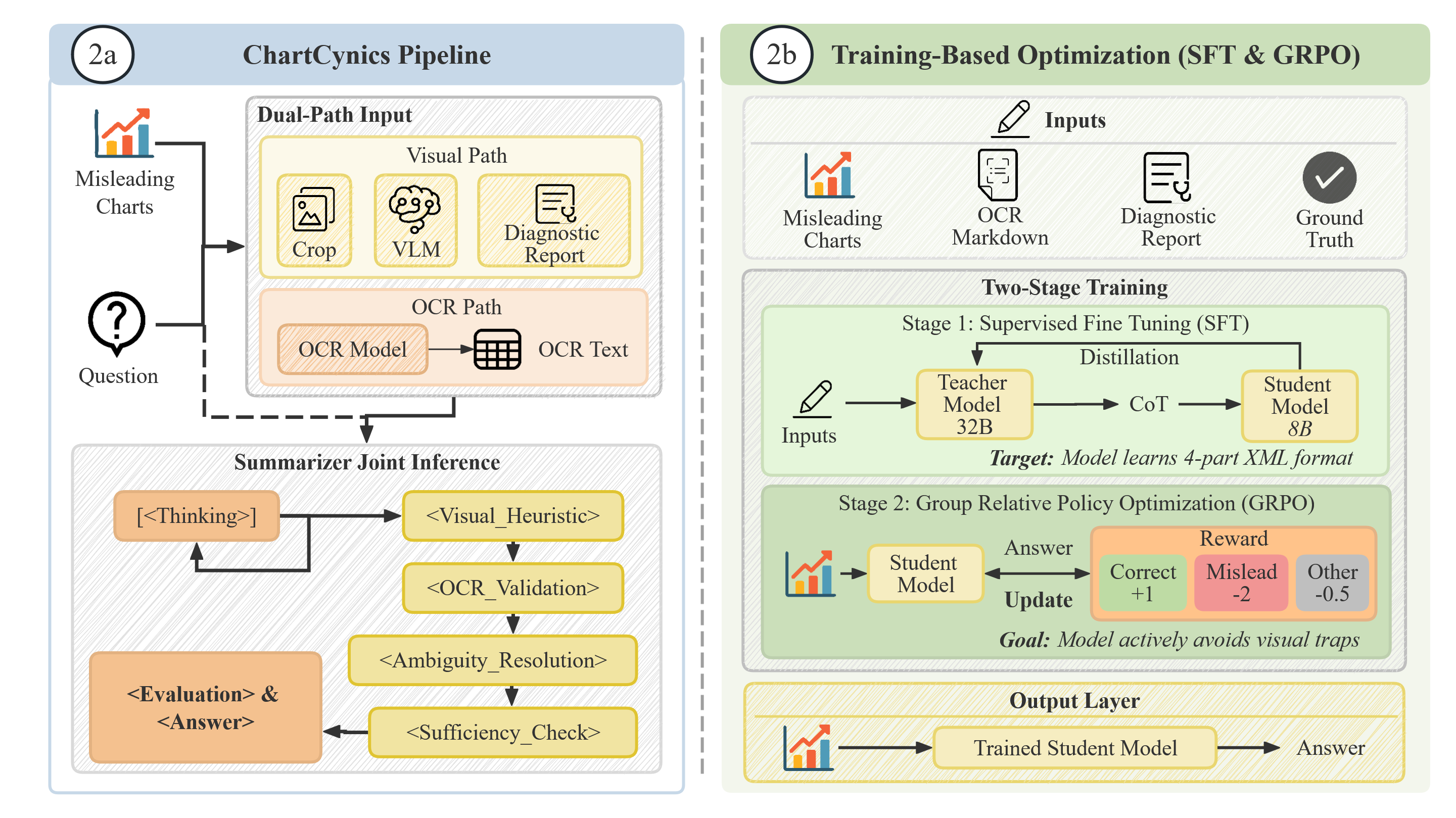}
\caption{Overview of the ChartCynics architecture. (a) Inference Pipeline: A training-free dual-path workflow that synthesizes visual diagnostics and OCR-extracted data through a structured reasoning chain. (b) Training-Based Optimization: A two-stage pipeline comprising Oracle-Informed SFT for logic distillation and Deception-Aware GRPO for adversarial alignment, utilizing asymmetric reward shaping to penalize visual traps.}
\label{fig:system_architecture}
\end{figure*}

To operationalize this objective and philosophy, we design a comprehensive architecture as illustrated in Figure \ref{fig:system_architecture}: the framework implements a dual-path pipeline that decouples visual heuristics from OCR-based numerical facts, feeding them into a \textit{Summarizer Joint Inference} module. To internalize this investigative logic, Figure \ref{fig:system_architecture}b details our two-stage optimization strategy: \textit{Oracle-Informed SFT} for reasoning distillation and \textit{Deception-Aware GRPO} for adversarial alignment. This unified architecture ensures that $a^*$ is the result of rigorous conflict resolution rather than visual sycophancy.

\subsection{Vision Path: Diagnostic-Augmented Investigation}

\begin{figure}[h]

    \centering

    \includegraphics[width=\linewidth]{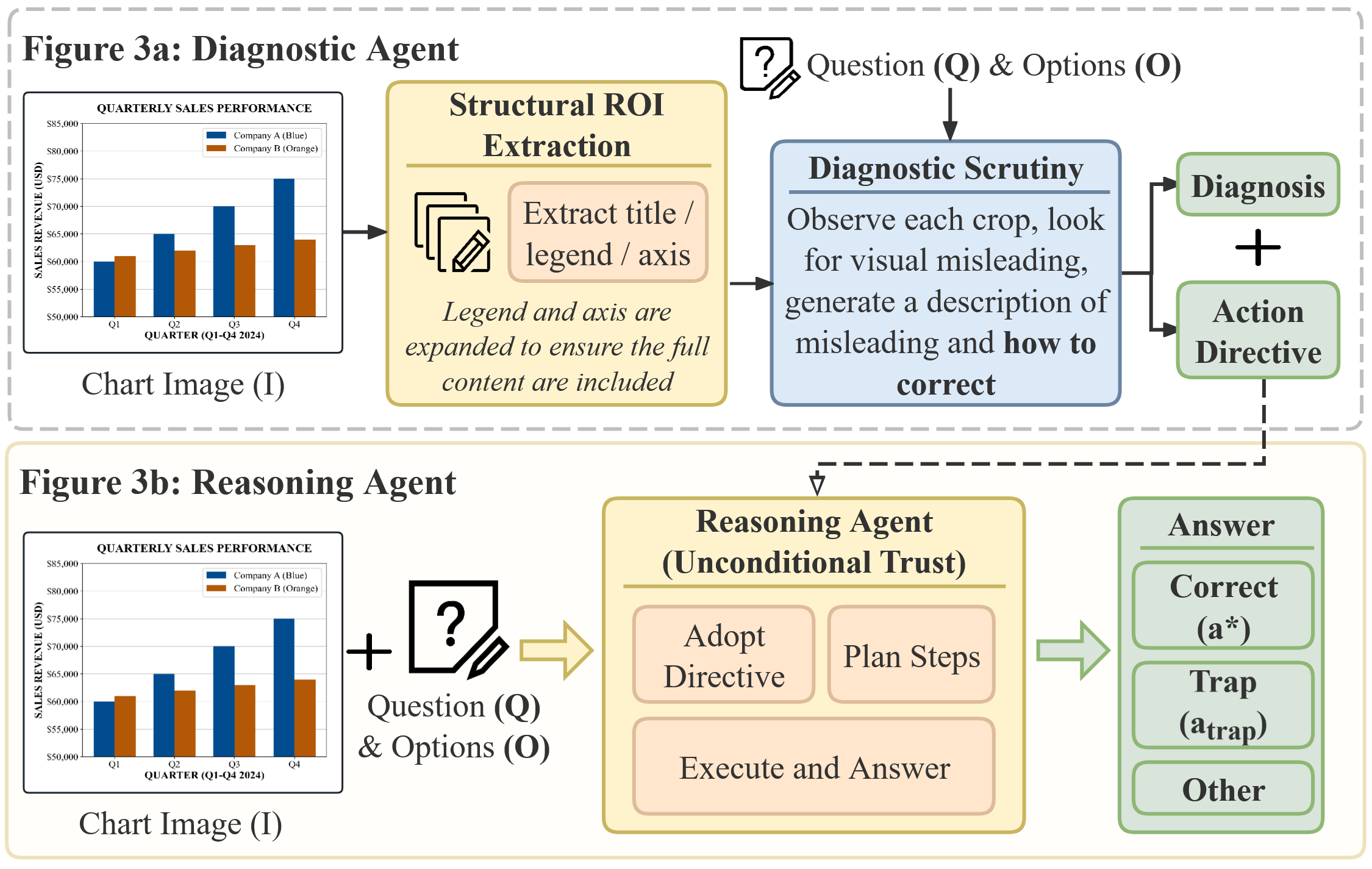}

    \caption{\textbf{Vision Path: Diagnostic-Augmented Investigation.} The framework decouples perception from reasoning to mitigate confirmation bias. The \textbf{Diagnostic Agent} identifies structural anomalies via high-resolution ROI extraction and generates an \textit{Action Directive}. Subsequently, the \textbf{Reasoning Agent} anchors its inference to these directives, ensuring that the final conclusion is grounded in structural evidence rather than global visual heuristics.}

    \label{fig:vision_path}

\end{figure}

As illustrated in Figure \ref{fig:vision_path}, the Vision Path operationalizes the primary visual component $P_v(I)$ of the agentic objective (Eq \ref{eq3}). 
Rather than a holistic glance, we decouple perception from reasoning through a two-agent architecture: the \textit{Diagnostic Agent} ($f_{diag}$) and the \textit{Reasoning Agent} ($f_{reason}$). The goal of this path is to generate a visual-perspective report that identifies potential traps before they reach the final fusion stage.

\subsubsection{Structural ROI Extraction and Semantic Padding}
To capture fine-grained deceptive cues (e.g., tick labels or small legend markers), we implement an automated Region of Interest (ROI) extraction module. 
Unlike heuristic-based cropping, we utilize a graphic element detection module to precisely localize the bounding boxes of critical chart components, defining the set $\mathcal{C}_{roi}$ = \{\textit{title, legend, x\text{-}axis, y\text{-}axis}\}.

To ensure semantic completeness, we apply a padding mechanism. 
For each $c \in \mathcal{C}_{roi}$, the module calculates the spatial spread $(x\_spread, y\_spread)$ of detected elements. 
For instance, legend ROIs are expanded to ensure color-coded markers are included, while axis ROIs are padded vertically to encompass the full extent of tick labels. 
This ensures that the Diagnostic Agent performs literal ``reading'' rather than visual ``estimation'' of the scale.

\subsubsection{Decoupled Agentic Workflow.}
A core innovation of \textbf{ChartCynics} is the functional isolation designed to eliminate confirmation bias \cite{wan2025unveiling}.

As shown in Figure \ref{fig:vision_path}a, the Diagnostic Agent executes a \textit{Blind Test}. 
It is provided only with $\mathcal{I}$ and the high-resolution ROI crops, while $\mathcal{Q}$ and $\mathcal{O}$ are strictly withheld. 
This forces the agent to act as a neutral auditor:
\begin{equation}
    \mathcal{R}_{diag} = f_{diag}(\text{Crops}(\mathcal{I}, \mathcal{C}_{roi}), \mathcal{T})
\end{equation}
where $\mathcal{R}_{diag}$ is the resulting diagnostic Report. 
This prevents ``Self-Prompt Pollution'' where the model might otherwise hallucinate visual evidence to support a pre-conceived answer derived from the question \cite{leng2024mitigating,zhou2023analyzing}.

The agent sequentially examines each crop for visual anomalies defined in $\mathcal{T}$. 
By amplifying these localized regions, the process uncovers deceptive structural cues that are otherwise masked by the intuitive global visual trend $\mathcal{V}_{trend}$.

\subsubsection{Cognitive Anchoring and Inference}
The Vision Path produces in a structured report $\mathcal{R}_{diag} = \{\text{Diagnosis}, \text{Action Directive}\}$. 
The \textit{Diagnosis} provides factual identification (e.g., ``The y-axis starts at 15,000''), while the \textit{Action Directive} provides prescriptive logic (e.g., ``Ignore visual heights; read literal tick values'').

To ensure objective inference, the \textbf{Reasoning Agent} (Figure \ref{fig:vision_path}b) follows an unconditional trust policy. 
It is required to explicitly anchor its Chain-of-Thought (CoT) to $\mathcal{R}_{diag}$ in its first reasoning step:
\begin{equation}
    a^* = f_{reason}(\mathcal{Q}, \mathcal{O}, \mathcal{R}_{diag})
\end{equation}

By forcing the model to reiterate the Action Directive, we shift the reasoning trajectory from an uncalibrated visual prior $P(a_{trap} \mid \mathcal{V}_{trend})$ to a comprehensive posterior $P(a^* \mid \mathcal{V}_{trend}, \mathcal{D}_{rel}, \mathcal{T})$ that fuses both modalities.
This mechanism effectively transforms the ``Trap Answer'' into a detectable anomaly, leading to the final correct inference $a^*$.

\subsection{Data Path: OCR-Driven Serialization}


To provide an orthogonal verification layer against visual illusions $\mathcal{V}_{trend}$, the Data Path ($\mathcal{P}_d$) is engineered to reconstruct the underlying numerical relationship. This path bypasses deceptive visual encodings (e.g., manipulated areas, lengths, or angles) by establishing a ``literal backbone'' through structured text and numeric data extraction.

\subsubsection{Structural Data Serialization}
Rather than relying on visual rendering, the first stage directly reveals the graphical attributes of the chart image $\mathcal{I}$. 
We utilize an advanced multimodal OCR parsing module to extract all explicit textual and numerical entities, such as axis scales, data labels, and legends, and serialize them into a unified, structured Markdown format $\mathcal{M}_{ocr}$:
\begin{equation}
    \mathcal{M}_{ocr} = f_{extract}(\mathcal{I})
\end{equation}

As this Markdown representation relies strictly on the parsed literal characters rather than their spatial positioning, it effectively recovers the numerical data $\mathcal{D}$ without being tainted by the deceptive visual manipulation $f_{deceptive}$ applied to the chart.

\subsubsection{Calibration Directives for Downstream Reasoning}
While $\mathcal{M}_{ocr}$ provides objective numerical evidence, optical extraction can sometimes capture numbers from arbitrary text or hallucinate when labels are absent. 
Therefore, rather than hard-coding a pre-processing filter, the Data Path establishes a set of \textbf{Calibration Directives} that are passed to the downstream Reasoning Agent to evaluate $\mathcal{M}_{ocr}$ dynamically during its Chain-of-Thought (CoT).

\begin{itemize}[leftmargin=*]
    \item \textbf{Dynamic Trust Evaluation:} Not all extracted values in $\mathcal{M}_{ocr}$ hold equal epistemic weight. 
    The framework defines a dual-tier trust principle: (1) \textit{High Trust} applies to values originating from explicit data labels plotted directly on elements, treating them as immutable ground truth. 
    (2) \textit{Low Trust (Skepticism)} applies when the chart relies solely on axis ticks without direct labels, or when visual elements exceed the canvas boundary. 
    In such cases, the agent is instructed to treat the OCR data as potentially imprecise and rely more heavily on visual diagnostic deductions.
    
    \item \textbf{Temporal and Structural Integrity Check:} To counter temporal deceptions like \textit{Inappropriate Aggregation} or \textit{Inappropriate Ordering}, the Data Path mandates an integrity audit rule. 
    The agent is explicitly instructed to scrutinize the sequence of categories in $\mathcal{M}_{ocr}$. 
    For instance, it must check whether the time-series is deliberately reversed, or if the final data point represents a deceptive ``incomplete period'' masquerading as a full cycle. 
\end{itemize}

\subsection{Agentic Fusion with Detective Chain-of-Thought}
The Agentic Fusion module acts as the \textbf{Summarizer} ($\mathcal{S}$), tasked with integrating the Diagnostic Report $\mathcal{R}_{diag}$ from the Vision Path ($\mathcal{P}_v$) and the Calibrated Data $\mathcal{M}_{ocr}$ from the Data Path ($\mathcal{P}_d$). 
As defined in the task formalization, its objective is to maximize the posterior $P(a \mid \mathcal{P}_v, \mathcal{P}_d, \mathcal{Q}, \mathcal{T})$ by explicitly identifying and neutralizing the manipulation function $f_{deceptive}$. Before detailing the training process, we first mathematically formalize the ideal reasoning policy that the Summarizer must follow via structured prompting to achieve this objective.

\subsubsection{Conflict Arbitration via Evidence Weighting}To resolve discrepancies between visual heuristics and numerical literals, the Summarizer $\mathcal{S}$ implements a hierarchical weight system governed by two \textbf{Golden Rules of Evidence}:
\begin{itemize}[leftmargin=*]
\item \textbf{Rule I: Heuristic Calibration.} Rather than a naive zero-sum arbitration that bluntly discards information from one conflicting path, the Summarizer recontextualizes it using the structural anomaly $\tau \in \mathcal{R}_{diag}$. Formally, the reasoning prompt acts as a structural constraint, forcing the agent's generation trajectory to align the visually-derived inference with the data-derived truth. The agent recognizes that the calibrated visual posterior leads to the same optimal decision as the numerical data:
\begin{equation}
\arg\max_{a \in \mathcal{O}} P(a \mid \mathcal{V}_{trend}, \tau) = \arg\max_{a \in \mathcal{O}} P(a \mid \mathcal{D}_{rel}) = a^*
\end{equation}
This ensures the model logically translates the visual shape (e.g., recognizing that a physical downward slope $\mathcal{V}_{trend}$ on an inverted axis $\tau$ perfectly corroborates a numerical increase derived from $\mathcal{D}_{rel}$ in Figure~\ref{fig:framework_case}).

\item \textbf{Rule II: Dynamic Trust Calibration.} The epistemic weight of $\mathcal{M}_{ocr}$ is conditioned on the \textit{Trust Level} flags. 
If $\mathcal{M}_{ocr}$ contains \textit{High Trust} labels, it serves as the immutable ground truth for $D_{rel}$. Conversely, if labels are absent and $\mathcal{R}_{diag}$ indicates a scale violation (e.g., ``Exceeding the Canvas''), the agent treats $\mathcal{M}_{ocr}$ as a qualitative indicator and prioritizes the \textit{Action Directives} for relative reasoning.
\end{itemize}

\subsubsection{The Five-Step Detective Framework (D-CoT)}
The reasoning process is operationalized via a structured Detective Chain-of-Thought (D-CoT), designed to transform the model from a passive describer into a skeptical auditor. 
This sequence ensures that the final selection $a^*$ is a logical consequence of debunking the $f_{deceptive}$ mechanism:
\begin{enumerate}[leftmargin=*]
\item \textbf{Perception Audit (Prior Calibration):} The agent first reiterates the \textit{Action Directives} from $\mathcal{R}_{diag}$. This step acts as a cognitive anchor to re-evaluate (rather than strictly suppress) the high-confidence prior $P(a_{trap} \mid \mathcal{V}_{trend})$ before performing any calculations.

\item \textbf{Numerical Anchoring ($D_{rel}$ Reconstruction):} The agent maps categories and values from $\mathcal{M}_{ocr}$ to the entities in $\mathcal{Q}$. It identifies whether the data represents the ``actual relationship'' $\mathcal{D}_{rel}$ or just a partial subset of the visual representation.

\item \textbf{Deception Mapping:} Using the Taxonomy $\mathcal{T}$ as a logic gate, the agent classifies the chart's specific manipulation subtype. For example, if mapped to \textit{Inappropriate Ordering}, the agent is triggered to re-sort the $\mathcal{M}_{ocr}$ sequence before comparison.

\item \textbf{Sufficiency \& Integrity Check:} The agent evaluates if the combined evidence ($\mathcal{R}_{diag} + \mathcal{M}_{ocr}$) is sufficient to resolve the contradiction. If the trust levels are conflicting or inadequate, it is permitted to output a ``Cannot be Inferred'' conclusion to avoid hallucination.

\item \textbf{Adversarial Trap Rejection:} In the final step, the agent must explicitly prove why the ``Trap Answer'' $a_{trap}$ (derived from $\mathcal{V}_{trend}$) is a mathematical fallacy. By articulating the discrepancy between the visual illusion and the grounded data, the model ensures $a^*$ is selected through critical elimination rather than pattern matching.
\end{enumerate}

\begin{figure}[htbp]
    \centering
    \includegraphics[width=\linewidth]{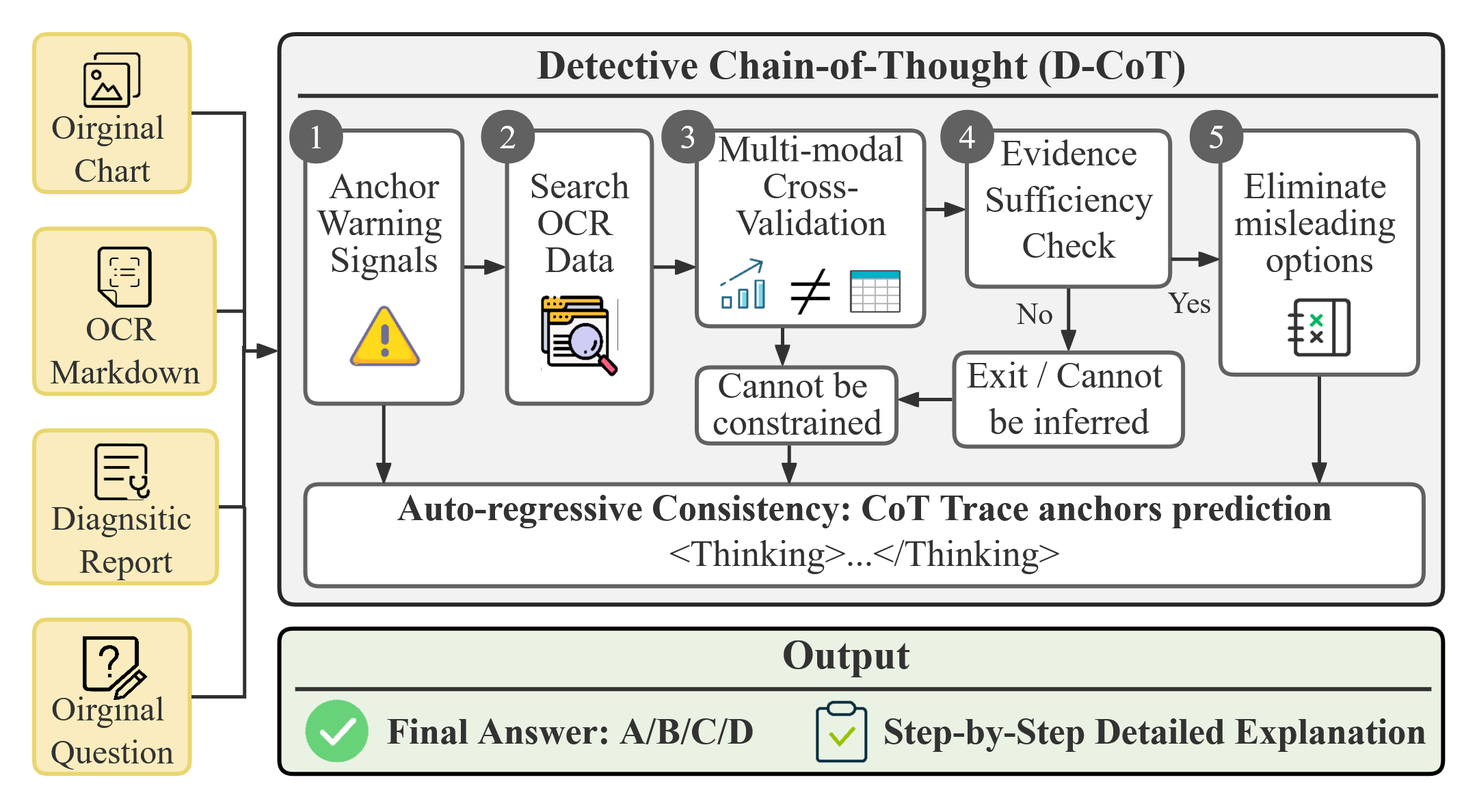} 
    \caption{The Agentic Fusion with Detective Chain-of-Thought (D-CoT) process. The Reasoning Agent integrates the Diagnostic Report and OCR Markdown through a five-step detective framework, utilizing Golden Rules and Misleader Taxonomy to ensure adversarial robustness and evidence-based deduction.}
    \label{fig:agentic_fusion}
\end{figure}

\subsection{Optimization: SFT and GRPO}

To internalize the reasoning heuristics and resolve multimodal conflicts, we implement a two-stage optimization pipeline: Oracle-Informed Distillation followed by Deception-Aware Group Relative Policy Optimization (GRPO).


\subsubsection{SFT via Oracle-Informed Reasoning Distillation}

We perform Supervised Fine-Tuning (SFT) using a dataset generated by a larger model (Qwen3-VL-32B).
To address the information disparity between the source model (which has access to raw CSV data) and the target model (which only processes visual pixels), we employ a specific prompting strategy.
During data generation, the larger model utilizes ground-truth CSV data and trap labels to ensure factual accuracy.
However, it is strictly constrained to generate reasoning chains based only on visible chart elements.
This process converts inaccessible structured data into visually-grounded logical traces, including $\langle Visual\_Heuristic \rangle$, $\langle OCR\_Validation \rangle$, and $\langle Ambiguity\_Resolution \rangle$.
Consequently, the target model learns to derive conclusions from evidence present within the image rather than relying on external metadata.

\subsubsection{Deception-Aware GRPO}
To evolve the model beyond simple imitation, we apply Group Relative Policy Optimization (GRPO)~\cite{shao2024deepseekmath} to align its reasoning process with adversarial skepticism. 
Unlike DPO~\cite{rafailov2023direct}, which relies on binary preferences, GRPO allows us to utilize a multi-dimensional continuous reward function $R_{total}$ to evaluate the quality of the generated reasoning trace within a sampled group $G$:\begin{equation}R_{total} = w_1 R_{fact} + w_2 R_{contra} + w_3 R_{logic} + w_4 R_{fmt} + R_{shaping}\end{equation}where the components are defined as follows:
\begin{itemize}[leftmargin=*]
\item \textbf{Numerical Grounding ($R_{fact}$):} To ensure robustness against OCR noise, $R_{fact}$ calculates the \textit{Spearman rank correlation} between numeric tokens extracted from the $\langle OCR\_Validation \rangle$ tag and the Oracle CSV. 
By rewarding the correlation of trends (rank) rather than absolute pixel-to-value accuracy, we encourage the model to prioritize relative data relationships.
\item \textbf{Semantic Contradiction ($R_{contra}$):} To prevent ``reward hacking'' via keyword stuffing, $R_{contra}$ combines keyword hit rates (e.g., ``non-zero baseline'') with a \textit{semantic overlap constraint}. The model is rewarded only if its identified contradiction aligns with the expert-annotated explanation, ensuring logical consistency across the 5-step D-CoT trajectory.
\item \textbf{Structural Enforcement ($R_{fmt}$):} A binary gatekeeper reward is applied to penalize ``shortcut'' generations. 
If the output fails to strictly follow the four-step XML-tagged structure, it incurs a significant baseline deduction, leveraging the LLM's auto-regressive nature to anchor the final answer in the preceding reasoning steps.
\item \textbf{Asymmetric Shaping ($R_{shaping}$):} We implement an adversarial penalty to combat visual sycophancy. 
While a correct answer yields $+1.0$, selecting the specific \textit{misleading distractor} (the trap) triggers a heavy penalty of $-2.0$. 
This asymmetry forces the policy to prioritize ``trap avoidance'' as a primary objective, effectively training the model to be inherently skeptical of deceptive visual cues.
\end{itemize}
By optimizing for the group-relative advantage of these fine-grained rewards, the model learns to navigate the tension between visual intuition and factual evidence without the computational overhead of a dedicated Critic network.
\section{Experiments}

This section evaluates the efficacy of \textbf{ChartCynics} in mitigating visual misleading. We detail the experimental configuration, followed by a mechanistic analysis of performance gains, ablation experiments, and the impact of our specialized RL alignment.

\begin{table*}[t]
\centering
\caption{Comparison between state-of-the-art baselines and the ChartCynics framework. We report Accuracy (Acc), Wrong due to Misleader (WM), and Wrong due to Others (WO). Values in bold indicate the best performance.}
\label{tab:main_results} 
\begin{tabular}{lcccccc}
\toprule
 & \multicolumn{3}{c}{\textbf{MC ($N=305$)}} & \multicolumn{3}{c}{\textbf{CDCC ($N=110$)}} \\ 
 \cmidrule(lr){2-4} \cmidrule(lr){5-7}
\textbf{Model \& Configuration} & \textbf{Acc $\uparrow$} & \textbf{WM $\downarrow$} & \textbf{WO $\downarrow$} & \textbf{Acc $\uparrow$} & \textbf{WM $\downarrow$} & \textbf{WO $\downarrow$} \\ \midrule
\rowcolor[gray]{0.95} \textit{Standard Multimodal Baselines} & & & & & & \\
ChartMoE \cite{xu2024chartmoe} & 33.11 & 43.28 & 23.61 & 28.18 & 46.36 & 25.45 \\ \midrule
\rowcolor[gray]{0.95} \textit{Structural Mitigation Heuristics} & & & & & & \\
Qwen3-VL-8B + Table-based QA \cite{tonglet2025protecting} & 49.18 & 29.51 & 21.31 & 36.39 & 20.91 & 42.73 \\
Qwen3-VL-8B + Redrawn \cite{tonglet2025protecting} & 44.26 & 35.41 & 20.33 & 38.18 & 20.91 & 40.91 \\ \midrule
\rowcolor[gray]{0.95} \textit{Ours: ChartCynics Framework over Base Models} & & & & & & \\
Qwen3-VL-8B (Base Model)~\cite{bai2025qwen3} & 45.57 & 40.00 & 14.43 & 35.45 & 31.82 & 32.73 \\
\quad + \textbf{ChartCynics} (Train-free) & 60.66 & 24.26 & 15.08 & 47.47 & 16.16 & 36.36 \\
\quad + \textbf{SFT + GRPO (Full)} & 74.43 & 11.15 & 14.42 & 64.55 & 8.18 & 27.27 \\ \addlinespace
o4-mini (Base Model)~\cite{openai2025o3o4mini} & 55.08 & 34.43 & 10.49 & 69.09 & 17.27 & 13.64 \\
\quad + \textbf{ChartCynics} (Train-free) & 71.80 & 18.69 & 9.51 & 79.09 & 9.09 & 11.82 \\ \addlinespace
Gemini-2.5-Flash (Base Model)~\cite{team2023gemini} & 67.21 & 20.98 & 11.80 & 62.73 & 17.27 & 20.00 \\
\quad + \textbf{ChartCynics} (Train-free) & 71.67 & 13.67 & 14.67 & 75.45 & 7.27 & 17.27 \\ \addlinespace
Gemini-3.1-Pro (Base Model)~\cite{gemini31pro} & 70.49 & 19.34 & 10.16 & 68.18 & 14.54 & 17.27 \\
\quad + \textbf{ChartCynics} (Train-free) & \textbf{81.31} & \textbf{9.51} & \textbf{9.18} & \textbf{86.36} & \textbf{2.73} & \textbf{10.91} \\ 
\bottomrule
\end{tabular}
\end{table*}

\subsection{Datasets and Evaluation Benchmarks}

To assess our framework's robustness and generalization, we evaluate ChartCynics across three distinct benchmarks. Detailed construction protocols, sampling strategies, and statistics for all datasets are presented in the supplementary material.

\begin{itemize}[leftmargin=*]
\item \textbf{Misleading ChartQA (MC) \cite{chen2025unmasking}:} Our primary benchmark for structural visual deception. We utilize the training set (2,619 samples) for our SFT and GRPO optimization, and evaluate on the official test set (305 samples).
\item \textbf{Curated Deceptive Chart Collection (CDCC) \cite{tonglet2025protecting}:} To test resilience against real-world epistemic conflicts, we compile a secondary benchmark of 110 expert-validated deceptive visualizations aggregated from established studies in the HCI community~\cite{ge2023calvi, lauer2020deceptive, lo2022misinformed}.
\item \textbf{Mixed Standard and Misleading Benchmark (MSMB) \cite{wu2024chartinsights, chen2025unmasking}:} To ensure our ``skeptical auditor'' does not suffer from over-skepticism on benign data, we construct a balanced evaluation set of 244 charts (122 standard \cite{wu2024chartinsights}, 122 misleading \cite{chen2025unmasking}) to verify fundamental chart comprehension remains intact.
\end{itemize}

\subsection{Metrics}
To evaluate both authentic reasoning capabilities and specific vulnerabilities to visual traps, we adopt three core metrics introduced in \cite{chen2025unmasking}. Formal definitions and detailed error attribution criteria are available in the supplementary material.

\begin{itemize}[leftmargin=*]
\item \textbf{Accuracy (Acc):} The standard grounded accuracy indicating the percentage of correctly answered queries.
\item \textbf{WM (Wrong due to Misleader):} A targeted metric tracking errors strictly caused by cognitive sycophancy, where the model's prediction falls for the visual trap.
\item \textbf{WO (Wrong due to Other factors):} A general error metric tracking failures unrelated to visual deception (e.g., basic OCR failures or instruction-following errors).
\end{itemize}

\subsection{Baselines}

To evaluate \textbf{ChartCynics}, we benchmark against three categories of state-of-the-art methods representing the paradigms discussed in our Introduction. First, to represent the visual-first paradigm, we evaluate proprietary VLMs (o4-mini~\cite{openai2025o3o4mini}, Gemini-2.5-Flash, and Gemini-3.1-Pro ~\cite{gemini31pro}) and our foundational backbone, Qwen3-VL-8B~\cite{bai2025qwen3}, to directly quantify the net gains of our dual-path architecture. Second, we include ChartMoE~\cite{xu2024chartmoe} as a domain-specific baseline to test if standard chart proficiency yields adversarial robustness. Finally, representing the data-first (OCR-enhanced) paradigm, we compare against two inference-time structural mitigations from \cite{tonglet2025protecting}: \textit{Table-based QA} and \textit{Visualization Redrawing}. This contrast effectively highlights the superiority of our synergistic arbitration over rigid, single-path de-rendering techniques.

\subsection{Implementation Details}

The training and evaluation of \textbf{ChartCynics} are conducted on a cluster of 4$\times$ NVIDIA A800 (80GB) GPUs. 
For dual-path input processing, we utilize \texttt{nemotron-graphic-elements-v1}~\cite{nemotron_graphic_elements_v1} for precise Region-of-Interest (ROI) bounding box extraction in the Vision Path. For the Data Path, we employ \texttt{LlamaParse}~\cite{llamacloud_docs} (configured with GPT-4o) as the underlying multimodal parsing engine—to robustly convert complex chart images into structured Markdown. We initialize our framework using the open-weight \texttt{Qwen3-VL-8B} as the foundational reasoning backbone and proceed with a two-stage optimization pipeline.

\textbf{Stage 1: Oracle-Informed SFT.} To bridge the epistemic gap between pixels and underlying data, we perform Supervised Fine-Tuning (SFT) on 5,238 high-quality reasoning chains. 
These chains are distilled from the 2,619 samples of the Misleading ChartQA training set \cite{chen2025unmasking} using a Qwen3-VL-32B teacher. 

\textbf{Stage 2: Deception-Aware GRPO.} Following SFT, we utilize GRPO for model alignment, employing a group size of $G=8$. The multi-objective reward function $R_{total}$ is defined with the following coefficients: $w_{fact}=0.20$, $w_{contra}=0.25$, $w_{logic}=0.20$, and $w_{fmt}=0.10$. Furthermore, we implement an asymmetric reward shaping strategy, applying a $-2.0$ penalty for selecting ``misleading trap'' distractors and a $+1.0$ reward for ground-truth alignment.

\subsection{Main Results}

To comprehensively evaluate our approach, Table \ref{tab:main_results} presents a detailed performance comparison across the MC and the CDCC. By analyzing the overall Accuracy (Acc) and the error breakdowns, we distill three primary observations:

\noindent \textbf{Superiority of Dual-Path Skepticism (Zero-Shot \& Aligned).}
To demonstrate broad applicability, we evaluate \textbf{ChartCynics} across backbones ranging from open-source to proprietary systems. The training-free framework shows immediate universality, boosting Qwen3-VL-8B by +15.09\% (to 60.66\%) and o4-mini by +16.72\% (to 71.80\%) on MC. Crucially, our two-stage optimization (SFT+GRPO) propels the 8B model to a state-of-the-art 74.43\% accuracy. This +28.86\% absolute gain enables the smaller model to outperform much larger proprietary systems, notably surpassing Gemini-3.1-Pro (70.49\%). This robust alignment further translates to the CDCC, where the optimized 8B model nearly doubles its baseline accuracy (35.45\% to 64.55\%).

\noindent \textbf{Mechanistic Suppression of Deception.}
The error breakdowns validate our Complementary Fusion hypothesis. 
Across all backbones, \textbf{ChartCynics} drastically reduces the WM rate without triggering a compensatory spike in the WO rate. 
For instance, the training-free o4-mini drops its WM rate nearly two-fold (from 34.43\% to 18.69\%) on MC. Most notably, the fully optimized Qwen3-VL-8B crushes its WM rate from 40.00\% down to 11.15\%. 
This confirms that the framework successfully anchors its reasoning in structurally-grounded OCR evidence rather than simply guessing blindly to avoid traps.

\noindent \textbf{Outperforming Structural Mitigation Heuristics.}
We investigate whether visual deception can be mitigated simply by removing the visual modality. As shown in the middle section of Table \ref{tab:main_results}, converting deceptive charts into raw tables (\textit{Table-based QA}) or standardizing them (\textit{Redrawn}) yields marginal gains or even degradation for Qwen3-VL-8B (+3.61\% and -1.31\%, respectively). This confirms that simply de-rendering an image into text is insufficient; models require the active, complementary reasoning provided by ChartCynics (74.43\%) to navigate complex structural deceptions.

Beyond deceptive scenarios, a critical concern is whether such defensive mechanisms degrade performance on benign data. We evaluate our training-free framework against ChartMoE \cite{xu2024chartmoe} on the Mixed Benchmark (Table \ref{tab:mixed_benchmark}). While ChartMoE degrades severely on visual deceptions (31.97\%), ChartCynics not only doubles this accuracy (68.03\%) but also surprisingly outperforms ChartMoE on standard charts (94.26\% vs. 88.52\%). The dual-agent structural investigation enhances fundamental data extraction across all chart types, proving our framework is a generalized solution rather than a narrow patch.

\begin{table}[h]
\centering
\caption{Performance on the Mixed Standard and Misleading Benchmark (MSMB).}
\label{tab:mixed_benchmark}
\begin{tabular}{lcc}
\toprule
\textbf{Data Subset} & \textbf{ChartMoE} & \textbf{ChartCynics} \\ 
 & \textbf{(SOTA)} & \textbf{(Train-free)} \\ \midrule
Standard ($N=122$) & 88.52\% (108) & \textbf{94.26\% (115)} \\
Misleading ($N=122$) & 31.97\% (39) & \textbf{68.03\% (83)} \\ \midrule
\textbf{Overall} ($N=244$) & 60.25\% (147) & \textbf{81.15\% (198)} \\ \bottomrule
\end{tabular}
\end{table}

\subsection{Ablation Study}

To isolate the contributions of our proposed framework, we conduct a comprehensive ablation study using MC (Table \ref{tab:combined_ablation}).

\noindent \textbf{Architectural Necessity of Complementary Fusion.} 
We first evaluate the necessity of merging visual structure with OCR literals. As shown in the upper section of Table \ref{tab:combined_ablation}, removing the visual diagnostic path (\textit{OCR Only}) yields the highest misled rate (\textbf{WM: 30.16\%}), confirming that unstructured OCR data lacks the spatial semantics required for accurate mapping. 
Conversely, enhancing the model with local crops (\textit{VLM + Crop}) effectively reduces observational errors (WO drops to 15.41\%) but fails to resolve the underlying deception. 
Only the full \textbf{ChartCynics} architecture successfully leverages multimodal complementarity to minimize both WM and WO simultaneously.

\noindent \textbf{Impact of Optimization Stages.} 
We further validate our two-stage optimization strategy (lower section of Table \ref{tab:combined_ablation}). 
Applying SFT alone provides a substantial performance leap (Acc: 68.52\%), distilling the investigative D-CoT logic into the model. 
Crucially, the subsequent GRPO alignment propels accuracy further to \textbf{74.43\%} and compresses the WM rate to a mere \textbf{11.15\%}. 
Notably, while GRPO drastically diminishes susceptibility to visual traps, it maintains a low rate of errors (WO: 14.42\%), proving that our reward shaping successfully aligns the model's cognitive circuits toward skeptical evidence-based reasoning rather than blind guessing.

\begin{table}[h]
\centering
\caption{Ablation Study (Qwen3-VL-8B) on MC.}
\label{tab:combined_ablation}
\begin{tabular}{lccc}
\toprule
\textbf{Setting} & \textbf{Acc $\uparrow$} & \textbf{WM $\downarrow$} & \textbf{WO $\downarrow$} \\ \midrule
\rowcolor[gray]{0.95} \textit{Architectural Components (Train-free)} & & & \\
VLM (Standard) & 52.13 & 27.21 & 20.66 \\
VLM + Crop & 57.38 & 27.21 & 15.41 \\
OCR Only (Vision-free) & 46.89 & 30.16 & 22.95 \\
\textbf{Full ChartCynics (Train-free)} & \textbf{60.66} & \textbf{24.26} & \textbf{15.08} \\ \midrule
\rowcolor[gray]{0.95} \textit{Optimization Stages (Full Architecture)} & & & \\
Full ChartCynics (Train-free) & 60.66 & 24.26 & 15.08 \\
+ SFT Only & 68.52 & 15.08 & 16.39 \\
+ \textbf{SFT + GRPO (Full)} & \textbf{74.43} & \textbf{11.15} & \textbf{14.42} \\ \bottomrule
\end{tabular}
\end{table}




\section{Conclusion}

This paper presents \textbf{ChartCynics}, an agentic dual-path framework that fortifies VLMs against misleading visualizations by decoupling intuitive perception from rigorous verification. By synergizing a \textit{Diagnostic Vision Path} with an \textit{OCR-Driven Data Path}, our approach systematically unmasks deceptive structures through a skeptical reasoning paradigm. Experimental results show that \textbf{ChartCynics} achieves a state-of-the-art \textbf{74.43\%} accuracy on the Qwen3-VL-8B backbone, outperforming advanced proprietary models. We demonstrate that while \textbf{SFT via Oracle-Informed Reasoning Distillation} is crucial for distilling structured investigative logic, the subsequent \textbf{Deception-Aware GRPO} is essential for penalizing visual sycophancy and enforcing logical consistency. Ultimately, ChartCynics provides a robust foundation for trustworthy multimedia AI, ensuring that automated data interpretation remains grounded in factual reality rather than deceptive visual heuristics.

\bibliographystyle{ACM-Reference-Format}
\bibliography{reference}


\begin{thebibliography}{55}


\ifx \showCODEN    \undefined \def \showCODEN     #1{\unskip}     \fi
\ifx \showISBNx    \undefined \def \showISBNx     #1{\unskip}     \fi
\ifx \showISBNxiii \undefined \def \showISBNxiii  #1{\unskip}     \fi
\ifx \showISSN     \undefined \def \showISSN      #1{\unskip}     \fi
\ifx \showLCCN     \undefined \def \showLCCN      #1{\unskip}     \fi
\ifx \shownote     \undefined \def \shownote      #1{#1}          \fi
\ifx \showarticletitle \undefined \def \showarticletitle #1{#1}   \fi
\ifx \showURL      \undefined \def \showURL       {\relax}        \fi
\providecommand\bibfield[2]{#2}
\providecommand\bibinfo[2]{#2}
\providecommand\natexlab[1]{#1}
\providecommand\showeprint[2][]{arXiv:#2}

\bibitem[Bai et~al\mbox{.}(2025)]%
        {bai2025qwen3}
\bibfield{author}{\bibinfo{person}{Shuai Bai}, \bibinfo{person}{Yuxuan Cai}, \bibinfo{person}{Ruizhe Chen}, \bibinfo{person}{Keqin Chen}, \bibinfo{person}{Xionghui Chen}, \bibinfo{person}{Zesen Cheng}, \bibinfo{person}{Lianghao Deng}, \bibinfo{person}{Wei Ding}, \bibinfo{person}{Chang Gao}, \bibinfo{person}{Chunjiang Ge}, {et~al\mbox{.}}} \bibinfo{year}{2025}\natexlab{}.
\newblock \showarticletitle{Qwen3-vl technical report}.
\newblock \bibinfo{journal}{\emph{arXiv preprint arXiv:2511.21631}} (\bibinfo{year}{2025}).
\newblock


\bibitem[Bendeck and Stasko(2024)]%
        {bendeck2024empirical}
\bibfield{author}{\bibinfo{person}{Alexander Bendeck} {and} \bibinfo{person}{John Stasko}.} \bibinfo{year}{2024}\natexlab{}.
\newblock \showarticletitle{An empirical evaluation of the GPT-4 multimodal language model on visualization literacy tasks}.
\newblock \bibinfo{journal}{\emph{IEEE Transactions on Visualization and Computer Graphics}} \bibinfo{volume}{31}, \bibinfo{number}{1} (\bibinfo{year}{2024}), \bibinfo{pages}{1105--1115}.
\newblock


\bibitem[Bharti et~al\mbox{.}(2024)]%
        {bharti2024chartom}
\bibfield{author}{\bibinfo{person}{Shubham Bharti}, \bibinfo{person}{Shiyun Cheng}, \bibinfo{person}{Jihyun Rho}, \bibinfo{person}{Jianrui Zhang}, \bibinfo{person}{Mu Cai}, \bibinfo{person}{Yong~Jae Lee}, \bibinfo{person}{Martina Rau}, {and} \bibinfo{person}{Xiaojin Zhu}.} \bibinfo{year}{2024}\natexlab{}.
\newblock \showarticletitle{CHARTOM: A Visual Theory-of-Mind Benchmark for LLMs on Misleading Charts}.
\newblock \bibinfo{journal}{\emph{arXiv preprint arXiv:2408.14419}} (\bibinfo{year}{2024}).
\newblock


\bibitem[Chaudhry et~al\mbox{.}(2020)]%
        {chaudhry2020leaf}
\bibfield{author}{\bibinfo{person}{Ritwick Chaudhry}, \bibinfo{person}{Sumit Shekhar}, \bibinfo{person}{Utkarsh Gupta}, \bibinfo{person}{Pranav Maneriker}, \bibinfo{person}{Prann Bansal}, {and} \bibinfo{person}{Ajay Joshi}.} \bibinfo{year}{2020}\natexlab{}.
\newblock \showarticletitle{Leaf-qa: Locate, encode \& attend for figure question answering}. In \bibinfo{booktitle}{\emph{Proceedings of the IEEE/CVF winter conference on applications of computer vision}}. \bibinfo{pages}{3512--3521}.
\newblock


\bibitem[Chen et~al\mbox{.}(2025a)]%
        {10.1145/3746027.3758288}
\bibfield{author}{\bibinfo{person}{Xiangnan Chen}, \bibinfo{person}{Yuancheng Fang}, \bibinfo{person}{Juncheng Li}, \bibinfo{person}{Qian Xiao}, \bibinfo{person}{Jun Lin}, \bibinfo{person}{Siliang Tang}, {and} \bibinfo{person}{Yueting Zhuang}.} \bibinfo{year}{2025}\natexlab{a}.
\newblock \showarticletitle{Chart-HQA: A Benchmark for Hypothetical Question Answering in Charts}. In \bibinfo{booktitle}{\emph{Proceedings of the 33rd ACM International Conference on Multimedia}} (Dublin, Ireland) \emph{(\bibinfo{series}{MM '25})}. \bibinfo{publisher}{Association for Computing Machinery}, \bibinfo{address}{New York, NY, USA}, \bibinfo{pages}{13297–13303}.
\newblock
\showISBNx{9798400720352}
\href{https://doi.org/10.1145/3746027.3758288}{doi:\nolinkurl{10.1145/3746027.3758288}}


\bibitem[Chen et~al\mbox{.}(2025b)]%
        {chen2025unmasking}
\bibfield{author}{\bibinfo{person}{Zixin Chen}, \bibinfo{person}{Sicheng Song}, \bibinfo{person}{Kashun Shum}, \bibinfo{person}{Yanna Lin}, \bibinfo{person}{Rui Sheng}, \bibinfo{person}{Weiqi Wang}, {and} \bibinfo{person}{Huamin Qu}.} \bibinfo{year}{2025}\natexlab{b}.
\newblock \showarticletitle{Unmasking deceptive visuals: Benchmarking multimodal large language models on misleading chart question answering}. In \bibinfo{booktitle}{\emph{Proceedings of the 2025 Conference on Empirical Methods in Natural Language Processing}}. \bibinfo{pages}{13767--13800}.
\newblock


\bibitem[Chowdhery et~al\mbox{.}(2023)]%
        {chowdhery2022palm}
\bibfield{author}{\bibinfo{person}{Aakanksha Chowdhery}, \bibinfo{person}{Sharan Narang}, \bibinfo{person}{Jacob Devlin}, \bibinfo{person}{Maarten Bosma}, \bibinfo{person}{Gaurav Mishra}, \bibinfo{person}{Adam Roberts}, \bibinfo{person}{Paul Barham}, \bibinfo{person}{Hyung~Won Chung}, \bibinfo{person}{Charles Sutton}, \bibinfo{person}{Sebastian Gehrmann}, {et~al\mbox{.}}} \bibinfo{year}{2023}\natexlab{}.
\newblock \showarticletitle{Palm: Scaling language modeling with pathways}.
\newblock \bibinfo{journal}{\emph{Journal of machine learning research}} \bibinfo{volume}{24}, \bibinfo{number}{240} (\bibinfo{year}{2023}), \bibinfo{pages}{1--113}.
\newblock


\bibitem[Comanici et~al\mbox{.}(2025)]%
        {team2023gemini}
\bibfield{author}{\bibinfo{person}{Gheorghe Comanici}, \bibinfo{person}{Eric Bieber}, \bibinfo{person}{Mike Schaekermann}, \bibinfo{person}{Ice Pasupat}, \bibinfo{person}{Noveen Sachdeva}, \bibinfo{person}{Inderjit Dhillon}, \bibinfo{person}{Marcel Blistein}, \bibinfo{person}{Ori Ram}, \bibinfo{person}{Dan Zhang}, \bibinfo{person}{Evan Rosen}, {et~al\mbox{.}}} \bibinfo{year}{2025}\natexlab{}.
\newblock \showarticletitle{Gemini 2.5: Pushing the frontier with advanced reasoning, multimodality, long context, and next generation agentic capabilities}.
\newblock \bibinfo{journal}{\emph{arXiv preprint arXiv:2507.06261}} (\bibinfo{year}{2025}).
\newblock


\bibitem[Correll and Heer(2017)]%
        {correll2017black}
\bibfield{author}{\bibinfo{person}{Michael Correll} {and} \bibinfo{person}{Jeffrey Heer}.} \bibinfo{year}{2017}\natexlab{}.
\newblock \showarticletitle{Black hat visualization}. In \bibinfo{booktitle}{\emph{Workshop on Dealing with Cognitive Biases in Visualisations (DECISIVe), IEEE VIS}}, Vol.~\bibinfo{volume}{1}. \bibinfo{pages}{10}.
\newblock


\bibitem[Dosovitskiy et~al\mbox{.}(2020)]%
        {dosovitskiy2020image}
\bibfield{author}{\bibinfo{person}{Alexey Dosovitskiy}, \bibinfo{person}{Lucas Beyer}, \bibinfo{person}{Alexander Kolesnikov}, \bibinfo{person}{Dirk Weissenborn}, \bibinfo{person}{Xiaohua Zhai}, \bibinfo{person}{Thomas Unterthiner}, \bibinfo{person}{Mostafa Dehghani}, \bibinfo{person}{Matthias Minderer}, \bibinfo{person}{Georg Heigold}, \bibinfo{person}{Sylvain Gelly}, {et~al\mbox{.}}} \bibinfo{year}{2020}\natexlab{}.
\newblock \showarticletitle{An image is worth 16x16 words: Transformers for image recognition at scale}.
\newblock \bibinfo{journal}{\emph{arXiv preprint arXiv:2010.11929}} (\bibinfo{year}{2020}).
\newblock


\bibitem[Engel({[n.\,d.]})]%
        {Engel}
\bibfield{author}{\bibinfo{person}{Pamela Engel}.} \bibinfo{year}{[n.\,d.]}\natexlab{}.
\newblock \bibinfo{title}{This chart shows an alarming rise in Florida gun deaths after ``Stand your ground`` was enacted}.
\newblock
\urldef\tempurl%
\url{https://www.businessinsider.com/gun-deaths-in-florida-increased-with-stand-your-ground-2014-2}
\showURL{%
\tempurl}


\bibitem[Ge et~al\mbox{.}(2023)]%
        {ge2023calvi}
\bibfield{author}{\bibinfo{person}{Lily~W Ge}, \bibinfo{person}{Yuan Cui}, {and} \bibinfo{person}{Matthew Kay}.} \bibinfo{year}{2023}\natexlab{}.
\newblock \showarticletitle{Calvi: Critical thinking assessment for literacy in visualizations}. In \bibinfo{booktitle}{\emph{Proceedings of the 2023 CHI conference on human factors in computing systems}}. \bibinfo{pages}{1--18}.
\newblock


\bibitem[{Google DeepMind}(2026)]%
        {gemini31pro}
\bibfield{author}{\bibinfo{person}{{Google DeepMind}}.} \bibinfo{year}{2026}\natexlab{}.
\newblock \bibinfo{title}{Gemini 3.1 Pro - Model Card}.
\newblock \bibinfo{howpublished}{\url{https://deepmind.google/models/model-cards/gemini-3-1-pro/}}.
\newblock
\newblock
\shownote{Accessed: 2026-05-29}.


\bibitem[Huff(2023)]%
        {huff2023how}
\bibfield{author}{\bibinfo{person}{Darrell Huff}.} \bibinfo{year}{2023}\natexlab{}.
\newblock \bibinfo{booktitle}{\emph{How to lie with statistics}}.
\newblock \bibinfo{publisher}{Penguin UK}.
\newblock


\bibitem[Kafle et~al\mbox{.}(2018)]%
        {kafle2018dvqa}
\bibfield{author}{\bibinfo{person}{Kushal Kafle}, \bibinfo{person}{Brian Price}, \bibinfo{person}{Scott Cohen}, {and} \bibinfo{person}{Christopher Kanan}.} \bibinfo{year}{2018}\natexlab{}.
\newblock \showarticletitle{Dvqa: Understanding data visualizations via question answering}. In \bibinfo{booktitle}{\emph{Proceedings of the IEEE conference on computer vision and pattern recognition}}. \bibinfo{pages}{5648--5656}.
\newblock


\bibitem[Kahou et~al\mbox{.}(2017)]%
        {kahou2017figureqa}
\bibfield{author}{\bibinfo{person}{Samira~Ebrahimi Kahou}, \bibinfo{person}{Vincent Michalski}, \bibinfo{person}{Adam Atkinson}, \bibinfo{person}{{\'A}kos K{\'a}d{\'a}r}, \bibinfo{person}{Adam Trischler}, {and} \bibinfo{person}{Yoshua Bengio}.} \bibinfo{year}{2017}\natexlab{}.
\newblock \showarticletitle{Figureqa: An annotated figure dataset for visual reasoning}.
\newblock \bibinfo{journal}{\emph{arXiv preprint arXiv:1710.07300}} (\bibinfo{year}{2017}).
\newblock


\bibitem[Lan and Liu(2024)]%
        {lan2024came}
\bibfield{author}{\bibinfo{person}{Xingyu Lan} {and} \bibinfo{person}{Yu Liu}.} \bibinfo{year}{2024}\natexlab{}.
\newblock \showarticletitle{“I Came Across a Junk”: Understanding Design Flaws of Data Visualization from the Public's Perspective}.
\newblock \bibinfo{journal}{\emph{IEEE transactions on visualization and computer graphics}} \bibinfo{volume}{31}, \bibinfo{number}{1} (\bibinfo{year}{2024}), \bibinfo{pages}{393--403}.
\newblock


\bibitem[Lauer and O'Brien(2020)]%
        {lauer2020deceptive}
\bibfield{author}{\bibinfo{person}{Claire Lauer} {and} \bibinfo{person}{Shaun O'Brien}.} \bibinfo{year}{2020}\natexlab{}.
\newblock \showarticletitle{The deceptive potential of common design tactics used in data visualizations}. In \bibinfo{booktitle}{\emph{Proceedings of the 38th ACM International Conference on Design of Communication}}. \bibinfo{pages}{1--9}.
\newblock


\bibitem[Lee et~al\mbox{.}(2023)]%
        {lee2023pix2struct}
\bibfield{author}{\bibinfo{person}{Kenton Lee}, \bibinfo{person}{Mandar Joshi}, \bibinfo{person}{Iulia~Raluca Turc}, \bibinfo{person}{Hexiang Hu}, \bibinfo{person}{Fangyu Liu}, \bibinfo{person}{Julian~Martin Eisenschlos}, \bibinfo{person}{Urvashi Khandelwal}, \bibinfo{person}{Peter Shaw}, \bibinfo{person}{Ming-Wei Chang}, {and} \bibinfo{person}{Kristina Toutanova}.} \bibinfo{year}{2023}\natexlab{}.
\newblock \showarticletitle{Pix2struct: Screenshot parsing as pretraining for visual language understanding}. In \bibinfo{booktitle}{\emph{International Conference on Machine Learning}}. PMLR, \bibinfo{pages}{18893--18912}.
\newblock


\bibitem[Lee et~al\mbox{.}(2016)]%
        {lee2016vlat}
\bibfield{author}{\bibinfo{person}{Sukwon Lee}, \bibinfo{person}{Sung-Hee Kim}, {and} \bibinfo{person}{Bum~Chul Kwon}.} \bibinfo{year}{2016}\natexlab{}.
\newblock \showarticletitle{Vlat: Development of a visualization literacy assessment test}.
\newblock \bibinfo{journal}{\emph{IEEE transactions on visualization and computer graphics}} \bibinfo{volume}{23}, \bibinfo{number}{1} (\bibinfo{year}{2016}), \bibinfo{pages}{551--560}.
\newblock


\bibitem[Leng et~al\mbox{.}(2024)]%
        {leng2024mitigating}
\bibfield{author}{\bibinfo{person}{Sicong Leng}, \bibinfo{person}{Hang Zhang}, \bibinfo{person}{Guanzheng Chen}, \bibinfo{person}{Xin Li}, \bibinfo{person}{Shijian Lu}, \bibinfo{person}{Chunyan Miao}, {and} \bibinfo{person}{Lidong Bing}.} \bibinfo{year}{2024}\natexlab{}.
\newblock \showarticletitle{Mitigating object hallucinations in large vision-language models through visual contrastive decoding}. In \bibinfo{booktitle}{\emph{Proceedings of the IEEE/CVF Conference on Computer Vision and Pattern Recognition}}. \bibinfo{pages}{13872--13882}.
\newblock


\bibitem[Liu et~al\mbox{.}(2023a)]%
        {liu2023deplot}
\bibfield{author}{\bibinfo{person}{Fangyu Liu}, \bibinfo{person}{Julian Eisenschlos}, \bibinfo{person}{Francesco Piccinno}, \bibinfo{person}{Syrine Krichene}, \bibinfo{person}{Chenxi Pang}, \bibinfo{person}{Kenton Lee}, \bibinfo{person}{Mandar Joshi}, \bibinfo{person}{Wenhu Chen}, \bibinfo{person}{Nigel Collier}, {and} \bibinfo{person}{Yasemin Altun}.} \bibinfo{year}{2023}\natexlab{a}.
\newblock \showarticletitle{DePlot: One-shot visual language reasoning by plot-to-table translation}. In \bibinfo{booktitle}{\emph{Findings of the Association for Computational Linguistics: ACL 2023}}. \bibinfo{pages}{10381--10399}.
\newblock


\bibitem[Liu et~al\mbox{.}(2023b)]%
        {liu2023matcha}
\bibfield{author}{\bibinfo{person}{Fangyu Liu}, \bibinfo{person}{Francesco Piccinno}, \bibinfo{person}{Syrine Krichene}, \bibinfo{person}{Chenxi Pang}, \bibinfo{person}{Kenton Lee}, \bibinfo{person}{Mandar Joshi}, \bibinfo{person}{Yasemin Altun}, \bibinfo{person}{Nigel Collier}, {and} \bibinfo{person}{Julian Eisenschlos}.} \bibinfo{year}{2023}\natexlab{b}.
\newblock \showarticletitle{Matcha: Enhancing visual language pretraining with math reasoning and chart derendering}. In \bibinfo{booktitle}{\emph{Proceedings of the 61st Annual Meeting of the Association for Computational Linguistics (Volume 1: Long Papers)}}. \bibinfo{pages}{12756--12770}.
\newblock


\bibitem[Liu and Chu(2024)]%
        {liu2024chart}
\bibfield{author}{\bibinfo{person}{Yi-Cheng Liu} {and} \bibinfo{person}{Wei-Ta Chu}.} \bibinfo{year}{2024}\natexlab{}.
\newblock \showarticletitle{Chart question answering based on modality conversion and large language models}. In \bibinfo{booktitle}{\emph{Proceedings of the 1st ACM Workshop on AI-Powered Q\&A Systems for Multimedia}}. \bibinfo{pages}{19--24}.
\newblock


\bibitem[LlamaIndex(2026)]%
        {llamacloud_docs}
\bibfield{author}{\bibinfo{person}{LlamaIndex}.} \bibinfo{year}{2026}\natexlab{}.
\newblock \bibinfo{title}{{LlamaCloud Documentation}}.
\newblock \bibinfo{howpublished}{\url{https://developers.llamaindex.ai/python/cloud/}}.
\newblock
\newblock
\shownote{Accessed: 2026-03-30}.


\bibitem[Lo et~al\mbox{.}(2022)]%
        {lo2022misinformed}
\bibfield{author}{\bibinfo{person}{Leo Yu-Ho Lo}, \bibinfo{person}{Ayush Gupta}, \bibinfo{person}{Kento Shigyo}, \bibinfo{person}{Aoyu Wu}, \bibinfo{person}{Enrico Bertini}, {and} \bibinfo{person}{Huamin Qu}.} \bibinfo{year}{2022}\natexlab{}.
\newblock \showarticletitle{Misinformed by visualization: What do we learn from misinformative visualizations?}. In \bibinfo{booktitle}{\emph{Computer Graphics Forum}}, Vol.~\bibinfo{volume}{41}. Wiley Online Library, \bibinfo{pages}{515--525}.
\newblock


\bibitem[Lo and Qu(2024)]%
        {lo2024good}
\bibfield{author}{\bibinfo{person}{Leo Yu-Ho Lo} {and} \bibinfo{person}{Huamin Qu}.} \bibinfo{year}{2024}\natexlab{}.
\newblock \showarticletitle{How good (or bad) are LLMs at detecting misleading visualizations?}
\newblock \bibinfo{journal}{\emph{IEEE Transactions on Visualization and Computer Graphics}} \bibinfo{volume}{31}, \bibinfo{number}{1} (\bibinfo{year}{2024}), \bibinfo{pages}{1116--1125}.
\newblock


\bibitem[Masry et~al\mbox{.}(2022)]%
        {masry2022chartqa}
\bibfield{author}{\bibinfo{person}{Ahmed Masry}, \bibinfo{person}{Xuan~Long Do}, \bibinfo{person}{Jia~Qing Tan}, \bibinfo{person}{Shafiq Joty}, {and} \bibinfo{person}{Enamul Hoque}.} \bibinfo{year}{2022}\natexlab{}.
\newblock \showarticletitle{Chartqa: A benchmark for question answering about charts with visual and logical reasoning}. In \bibinfo{booktitle}{\emph{Findings of the association for computational linguistics: ACL 2022}}. \bibinfo{pages}{2263--2279}.
\newblock


\bibitem[McNutt et~al\mbox{.}(2020)]%
        {mcnutt2020surfacing}
\bibfield{author}{\bibinfo{person}{Andrew McNutt}, \bibinfo{person}{Gordon Kindlmann}, {and} \bibinfo{person}{Michael Correll}.} \bibinfo{year}{2020}\natexlab{}.
\newblock \showarticletitle{Surfacing visualization mirages}. In \bibinfo{booktitle}{\emph{Proceedings of the 2020 CHI Conference on human factors in computing systems}}. \bibinfo{pages}{1--16}.
\newblock


\bibitem[Methani et~al\mbox{.}(2020)]%
        {methani2020plotqa}
\bibfield{author}{\bibinfo{person}{Nitesh Methani}, \bibinfo{person}{Pranay Ganguly}, \bibinfo{person}{Mitesh~M Khapra}, {and} \bibinfo{person}{Pratyush Kumar}.} \bibinfo{year}{2020}\natexlab{}.
\newblock \showarticletitle{PlotQA: Reasoning over Scientific Plots}. In \bibinfo{booktitle}{\emph{Proceedings of the IEEE/CVF Winter Conference on Applications of Computer Vision (WACV)}}.
\newblock


\bibitem[{NVIDIA}(2025)]%
        {nemotron_graphic_elements_v1}
\bibfield{author}{\bibinfo{person}{{NVIDIA}}.} \bibinfo{year}{2025}\natexlab{}.
\newblock \bibinfo{title}{Nemotron Graphic Elements v1}.
\newblock \bibinfo{howpublished}{\url{https://huggingface.co/nvidia/nemotron-graphic-elements-v1}}.
\newblock
\newblock
\shownote{Accessed: 2026-03-24}.


\bibitem[{OpenAI}(2025)]%
        {openai2025o3o4mini}
\bibfield{author}{\bibinfo{person}{{OpenAI}}.} \bibinfo{year}{2025}\natexlab{}.
\newblock \bibinfo{booktitle}{\emph{OpenAI o3 and o4-mini System Card}}.
\newblock \bibinfo{type}{{T}echnical {R}eport}. \bibinfo{institution}{OpenAI}.
\newblock
\urldef\tempurl%
\url{https://cdn.openai.com/pdf/2221c875-02dc-4789-800b-e7758f3722c1/o3-and-o4-mini-system-card.pdf}
\showURL{%
\tempurl}


\bibitem[Ouyang et~al\mbox{.}(2022)]%
        {ouyang2022training}
\bibfield{author}{\bibinfo{person}{Long Ouyang}, \bibinfo{person}{Jeffrey Wu}, \bibinfo{person}{Xu Jiang}, \bibinfo{person}{Diogo Almeida}, \bibinfo{person}{Carroll Wainwright}, \bibinfo{person}{Pamela Mishkin}, \bibinfo{person}{Chong Zhang}, \bibinfo{person}{Sandhini Agarwal}, \bibinfo{person}{Katarina Slama}, \bibinfo{person}{Alex Ray}, {et~al\mbox{.}}} \bibinfo{year}{2022}\natexlab{}.
\newblock \showarticletitle{Training language models to follow instructions with human feedback}.
\newblock \bibinfo{journal}{\emph{Advances in neural information processing systems}}  \bibinfo{volume}{35} (\bibinfo{year}{2022}), \bibinfo{pages}{27730--27744}.
\newblock


\bibitem[Pandey et~al\mbox{.}(2015)]%
        {pandey2015deceptive}
\bibfield{author}{\bibinfo{person}{Anshul~Vikram Pandey}, \bibinfo{person}{Katharina Rall}, \bibinfo{person}{Mar garet~L Satterthwaite}, \bibinfo{person}{Oded Nov}, {and} \bibinfo{person}{Enrico Bertini}.} \bibinfo{year}{2015}\natexlab{}.
\newblock \showarticletitle{How deceptive are deceptive visualizations? An empirical analysis of common distortion techniques}. In \bibinfo{booktitle}{\emph{Proceedings of the 33rd annual acm conference on human factors in computing systems}}. \bibinfo{pages}{1469--1478}.
\newblock


\bibitem[Pandey and Ottley(2025)]%
        {pandey2025benchmarking}
\bibfield{author}{\bibinfo{person}{Saugat Pandey} {and} \bibinfo{person}{Alvitta Ottley}.} \bibinfo{year}{2025}\natexlab{}.
\newblock \showarticletitle{Benchmarking visual language models on standardized visualization literacy tests}. In \bibinfo{booktitle}{\emph{Computer Graphics Forum}}, Vol.~\bibinfo{volume}{44}. Wiley Online Library, \bibinfo{pages}{e70137}.
\newblock


\bibitem[Rafailov et~al\mbox{.}(2023)]%
        {rafailov2023direct}
\bibfield{author}{\bibinfo{person}{Rafael Rafailov}, \bibinfo{person}{Archit Sharma}, \bibinfo{person}{Eric Mitchell}, \bibinfo{person}{Christopher~D Manning}, \bibinfo{person}{Stefano Ermon}, {and} \bibinfo{person}{Chelsea Finn}.} \bibinfo{year}{2023}\natexlab{}.
\newblock \showarticletitle{Direct preference optimization: Your language model is secretly a reward model}.
\newblock \bibinfo{journal}{\emph{Advances in neural information processing systems}}  \bibinfo{volume}{36} (\bibinfo{year}{2023}), \bibinfo{pages}{53728--53741}.
\newblock


\bibitem[Rho et~al\mbox{.}(2024)]%
        {rho2024various}
\bibfield{author}{\bibinfo{person}{Jihyun Rho}, \bibinfo{person}{Martina~A Rau}, \bibinfo{person}{Shubham~Kumar Bharti}, \bibinfo{person}{Rosanne Luu}, \bibinfo{person}{Jeremy McMahan}, \bibinfo{person}{Andrew Wang}, {and} \bibinfo{person}{Jerry Zhu}.} \bibinfo{year}{2024}\natexlab{}.
\newblock \showarticletitle{Various misleading visual features in misleading graphs: Do they truly deceive us?}. In \bibinfo{booktitle}{\emph{Proceedings of the Annual Meeting of the Cognitive Science Society}}, Vol.~\bibinfo{volume}{46}.
\newblock


\bibitem[Shao et~al\mbox{.}(2024)]%
        {shao2024deepseekmath}
\bibfield{author}{\bibinfo{person}{Zhihong Shao}, \bibinfo{person}{Peiyi Wang}, \bibinfo{person}{Qihao Zhu}, \bibinfo{person}{Runxin Xu}, \bibinfo{person}{Junxiao Song}, \bibinfo{person}{Xiao Bi}, \bibinfo{person}{Haowei Zhang}, \bibinfo{person}{Mingchuan Zhang}, \bibinfo{person}{YK Li}, \bibinfo{person}{Yang Wu}, {et~al\mbox{.}}} \bibinfo{year}{2024}\natexlab{}.
\newblock \showarticletitle{Deepseekmath: Pushing the limits of mathematical reasoning in open language models}.
\newblock \bibinfo{journal}{\emph{arXiv preprint arXiv:2402.03300}} (\bibinfo{year}{2024}).
\newblock


\bibitem[Szafir(2018)]%
        {szafir2018good}
\bibfield{author}{\bibinfo{person}{Danielle~Albers Szafir}.} \bibinfo{year}{2018}\natexlab{}.
\newblock \showarticletitle{The good, the bad, and the biased: Five ways visualizations can mislead (and how to fix them)}.
\newblock \bibinfo{journal}{\emph{interactions}} \bibinfo{volume}{25}, \bibinfo{number}{4} (\bibinfo{year}{2018}), \bibinfo{pages}{26--33}.
\newblock


\bibitem[Tonglet et~al\mbox{.}(2025)]%
        {tonglet2025protecting}
\bibfield{author}{\bibinfo{person}{Jonathan Tonglet}, \bibinfo{person}{Tinne Tuytelaars}, \bibinfo{person}{Marie-Francine Moens}, {and} \bibinfo{person}{Iryna Gurevych}.} \bibinfo{year}{2025}\natexlab{}.
\newblock \showarticletitle{Protecting multimodal large language models against misleading visualizations}.
\newblock \bibinfo{journal}{\emph{arXiv preprint arXiv:2502.20503}} (\bibinfo{year}{2025}).
\newblock


\bibitem[Tufte and Graves-Morris(1983)]%
        {tufte1983visual}
\bibfield{author}{\bibinfo{person}{Edward~R Tufte} {and} \bibinfo{person}{Peter~R Graves-Morris}.} \bibinfo{year}{1983}\natexlab{}.
\newblock \bibinfo{booktitle}{\emph{The visual display of quantitative information}}. Vol.~\bibinfo{volume}{2}.
\newblock \bibinfo{publisher}{Graphics press Cheshire, CT}.
\newblock


\bibitem[Valentim et~al\mbox{.}(2025)]%
        {valentim2025plot}
\bibfield{author}{\bibinfo{person}{Matheus Valentim}, \bibinfo{person}{Vaishali Dhanoa}, \bibinfo{person}{Gabriela~Molina Le{\'o}n}, {and} \bibinfo{person}{Niklas Elmqvist}.} \bibinfo{year}{2025}\natexlab{}.
\newblock \showarticletitle{The Plot Thickens: Quantitative Part-by-Part Exploration of MLLM Visualization Literacy}.
\newblock \bibinfo{journal}{\emph{arXiv preprint arXiv:2504.02217}} (\bibinfo{year}{2025}).
\newblock


\bibitem[Wan et~al\mbox{.}(2025)]%
        {wan2025unveiling}
\bibfield{author}{\bibinfo{person}{Yue Wan}, \bibinfo{person}{Xiaowei Jia}, {and} \bibinfo{person}{Xiang~Lorraine Li}.} \bibinfo{year}{2025}\natexlab{}.
\newblock \showarticletitle{Unveiling confirmation bias in chain-of-thought reasoning}. In \bibinfo{booktitle}{\emph{Findings of the Association for Computational Linguistics: ACL 2025}}. \bibinfo{pages}{3788--3804}.
\newblock


\bibitem[Wang et~al\mbox{.}(2026)]%
        {wang2025thinklite}
\bibfield{author}{\bibinfo{person}{Xiyao Wang}, \bibinfo{person}{Zhengyuan Yang}, \bibinfo{person}{Chao Feng}, \bibinfo{person}{Hongjin Lu}, \bibinfo{person}{Linjie Li}, \bibinfo{person}{Chung-Ching Lin}, \bibinfo{person}{Kevin Lin}, \bibinfo{person}{Furong Huang}, {and} \bibinfo{person}{Lijuan Wang}.} \bibinfo{year}{2026}\natexlab{}.
\newblock \showarticletitle{Sota with less: Mcts-guided sample selection for data-efficient visual reasoning self-improvement}.
\newblock \bibinfo{journal}{\emph{Advances in Neural Information Processing Systems}}  \bibinfo{volume}{38} (\bibinfo{year}{2026}), \bibinfo{pages}{118818--118850}.
\newblock


\bibitem[Wei et~al\mbox{.}(2022)]%
        {wei2022chain}
\bibfield{author}{\bibinfo{person}{Jason Wei}, \bibinfo{person}{Xuezhi Wang}, \bibinfo{person}{Dale Schuurmans}, \bibinfo{person}{Maarten Bosma}, \bibinfo{person}{Fei Xia}, \bibinfo{person}{Ed Chi}, \bibinfo{person}{Quoc~V Le}, \bibinfo{person}{Denny Zhou}, {et~al\mbox{.}}} \bibinfo{year}{2022}\natexlab{}.
\newblock \showarticletitle{Chain-of-thought prompting elicits reasoning in large language models}.
\newblock \bibinfo{journal}{\emph{Advances in neural information processing systems}}  \bibinfo{volume}{35} (\bibinfo{year}{2022}), \bibinfo{pages}{24824--24837}.
\newblock


\bibitem[Wei et~al\mbox{.}(2025)]%
        {wei2025chartmind}
\bibfield{author}{\bibinfo{person}{Jingxuan Wei}, \bibinfo{person}{Nan Xu}, \bibinfo{person}{Junnan Zhu}, \bibinfo{person}{Gaowei Wu}, \bibinfo{person}{Qi Chen}, \bibinfo{person}{Bihui Yu}, \bibinfo{person}{Lei Wang}, {et~al\mbox{.}}} \bibinfo{year}{2025}\natexlab{}.
\newblock \showarticletitle{Chartmind: A comprehensive benchmark for complex real-world multimodal chart question answering}. In \bibinfo{booktitle}{\emph{Proceedings of the 2025 Conference on Empirical Methods in Natural Language Processing}}. \bibinfo{pages}{4555--4569}.
\newblock


\bibitem[Wu et~al\mbox{.}(2024)]%
        {wu2024chartinsights}
\bibfield{author}{\bibinfo{person}{Yifan Wu}, \bibinfo{person}{Lutao Yan}, \bibinfo{person}{Leixian Shen}, \bibinfo{person}{Yunhai Wang}, \bibinfo{person}{Nan Tang}, {and} \bibinfo{person}{Yuyu Luo}.} \bibinfo{year}{2024}\natexlab{}.
\newblock \showarticletitle{Chartinsights: Evaluating multimodal large language models for low-level chart question answering}. In \bibinfo{booktitle}{\emph{Findings of the Association for Computational Linguistics: EMNLP 2024}}. \bibinfo{pages}{12174--12200}.
\newblock


\bibitem[Xia et~al\mbox{.}(2025)]%
        {xia2025chartx}
\bibfield{author}{\bibinfo{person}{Renqiu Xia}, \bibinfo{person}{Hancheng Ye}, \bibinfo{person}{Xiangchao Yan}, \bibinfo{person}{Qi Liu}, \bibinfo{person}{Hongbin Zhou}, \bibinfo{person}{Zijun Chen}, \bibinfo{person}{Botian Shi}, \bibinfo{person}{Junchi Yan}, {and} \bibinfo{person}{Bo Zhang}.} \bibinfo{year}{2025}\natexlab{}.
\newblock \showarticletitle{Chartx \& chartvlm: A versatile benchmark and foundation model for complicated chart reasoning}.
\newblock \bibinfo{journal}{\emph{IEEE Transactions on Image Processing}} (\bibinfo{year}{2025}).
\newblock


\bibitem[Xu et~al\mbox{.}(2025a)]%
        {xu2025chartpoint}
\bibfield{author}{\bibinfo{person}{Zhengzhuo Xu}, \bibinfo{person}{SiNan Du}, \bibinfo{person}{Yiyan Qi}, \bibinfo{person}{Siwen Lu}, \bibinfo{person}{Chengjin Xu}, \bibinfo{person}{Chun Yuan}, {and} \bibinfo{person}{Jian Guo}.} \bibinfo{year}{2025}\natexlab{a}.
\newblock \showarticletitle{Chartpoint: Guiding mllms with grounding reflection for chart reasoning}. In \bibinfo{booktitle}{\emph{Proceedings of the IEEE/CVF International Conference on Computer Vision}}. \bibinfo{pages}{426--436}.
\newblock


\bibitem[Xu et~al\mbox{.}(2025b)]%
        {xu2024chartmoe}
\bibfield{author}{\bibinfo{person}{Zhengzhuo Xu}, \bibinfo{person}{Bowen Qu}, \bibinfo{person}{Yiyan Qi}, \bibinfo{person}{Sinan Du}, \bibinfo{person}{Chengjin Xu}, \bibinfo{person}{Chun Yuan}, {and} \bibinfo{person}{Jian Guo}.} \bibinfo{year}{2025}\natexlab{b}.
\newblock \showarticletitle{Chartmoe: Mixture of diversely aligned expert connector for chart understanding}. In \bibinfo{booktitle}{\emph{International Conference on Learning Representations}}, Vol.~\bibinfo{volume}{2025}. \bibinfo{pages}{78550--78572}.
\newblock


\bibitem[Yang et~al\mbox{.}(2021)]%
        {yang2021truncating}
\bibfield{author}{\bibinfo{person}{Brenda~W Yang}, \bibinfo{person}{Camila~Vargas Restrepo}, \bibinfo{person}{Matthew~L Stanley}, {and} \bibinfo{person}{Elizabeth~J Marsh}.} \bibinfo{year}{2021}\natexlab{}.
\newblock \showarticletitle{Truncating bar graphs persistently misleads viewers}.
\newblock \bibinfo{journal}{\emph{Journal of Applied Research in Memory and Cognition}} \bibinfo{volume}{10}, \bibinfo{number}{2} (\bibinfo{year}{2021}), \bibinfo{pages}{298--311}.
\newblock


\bibitem[Yang et~al\mbox{.}(2024)]%
        {yang2024askchart}
\bibfield{author}{\bibinfo{person}{Xudong Yang}, \bibinfo{person}{Yifan Wu}, \bibinfo{person}{Yizhang Zhu}, \bibinfo{person}{Nan Tang}, {and} \bibinfo{person}{Yuyu Luo}.} \bibinfo{year}{2024}\natexlab{}.
\newblock \showarticletitle{Askchart: Universal chart understanding through textual enhancement}.
\newblock \bibinfo{journal}{\emph{arXiv preprint arXiv:2412.19146}} (\bibinfo{year}{2024}).
\newblock


\bibitem[Zhang et~al\mbox{.}(2023a)]%
        {zhang2023internlm}
\bibfield{author}{\bibinfo{person}{Pan Zhang}, \bibinfo{person}{Xiaoyi Dong}, \bibinfo{person}{Bin Wang}, \bibinfo{person}{Yuhang Cao}, \bibinfo{person}{Chao Xu}, \bibinfo{person}{Linke Ouyang}, \bibinfo{person}{Zhiyuan Zhao}, \bibinfo{person}{Haodong Duan}, \bibinfo{person}{Songyang Zhang}, \bibinfo{person}{Shuangrui Ding}, {et~al\mbox{.}}} \bibinfo{year}{2023}\natexlab{a}.
\newblock \showarticletitle{Internlm-xcomposer: A vision-language large model for advanced text-image comprehension and composition}.
\newblock \bibinfo{journal}{\emph{arXiv preprint arXiv:2309.15112}} (\bibinfo{year}{2023}).
\newblock


\bibitem[Zhang et~al\mbox{.}(2023b)]%
        {zhang2023multimodal}
\bibfield{author}{\bibinfo{person}{Zhuosheng Zhang}, \bibinfo{person}{Aston Zhang}, \bibinfo{person}{Mu Li}, \bibinfo{person}{Hai Zhao}, \bibinfo{person}{George Karypis}, {and} \bibinfo{person}{Alex Smola}.} \bibinfo{year}{2023}\natexlab{b}.
\newblock \showarticletitle{Multimodal chain-of-thought reasoning in language models}.
\newblock \bibinfo{journal}{\emph{arXiv preprint arXiv:2302.00923}} (\bibinfo{year}{2023}).
\newblock


\bibitem[Zhou et~al\mbox{.}(2024)]%
        {zhou2023analyzing}
\bibfield{author}{\bibinfo{person}{Yiyang Zhou}, \bibinfo{person}{Chenhang Cui}, \bibinfo{person}{Jaehong Yoon}, \bibinfo{person}{Linjun Zhang}, \bibinfo{person}{Zhun Deng}, \bibinfo{person}{Chelsea Finn}, \bibinfo{person}{Mohit Bansal}, {and} \bibinfo{person}{Huaxiu Yao}.} \bibinfo{year}{2024}\natexlab{}.
\newblock \showarticletitle{Analyzing and mitigating object hallucination in large vision-language models}. In \bibinfo{booktitle}{\emph{International Conference on Learning Representations}}, Vol.~\bibinfo{volume}{2024}. \bibinfo{pages}{56969--56998}.
\newblock


\end{thebibliography}

\end{document}